\documentclass[lettersize,journal]{IEEEtran}
\usepackage{amsmath,amsfonts}
\usepackage{algorithmic}
\usepackage{algorithm}
\usepackage{array}
\usepackage[caption=false,font=footnotesize,labelfont=rm,textfont=rm]{subfig}
\usepackage{textcomp}
\usepackage{stfloats}
\fnbelowfloat
\usepackage{url}
\usepackage{verbatim}
\usepackage{graphicx}
\usepackage{cite}
\usepackage{multirow}
\usepackage{etoolbox}
\usepackage{pifont}
\usepackage{xcolor}
\usepackage[colorlinks,
            linkcolor=black,
            anchorcolor=black,
            urlcolor=black,
            citecolor=black]{hyperref}
\usepackage{booktabs}
\usepackage{tabularx}  
\usepackage{caption}
\usepackage{cases}
\usepackage{float}
\usepackage{xspace}
\usepackage{amssymb}

\definecolor{deepblue}{RGB}{0, 31, 175}

\newcommand{\ACRONYM}{FLAP\xspace{}}

\begin{document}
\setlength{\skip\footins}{12pt}

\title{FLAP: FOV-Constrained Active Perception Planning for Prior-Map-Free 3D Navigation}
\author{Mengke Zhang\textsuperscript{1,2}, Sitong Li\textsuperscript{2}, Tiancheng Lai\textsuperscript{1,2}, Ruitian Pang\textsuperscript{1,2}, Mingxuan Zhang\textsuperscript{1,2}, Qingcheng Chen\textsuperscript{3}, Fei Gao\textsuperscript{1}, Chao Xu\textsuperscript{1,2}, Yanjun Cao\textsuperscript{1,2}
        \thanks{$^*$This work was supported by the Key R\&D Project of China National Tobacco Corporation under Grant No. 110202402018. }
        \thanks{$^1$The State Key Laboratory of Industrial Control Technology, College of Control Science and Engineering, Zhejiang University, Hangzhou 310027, China.}
        \thanks{$^2$Huzhou Institute, Zhejiang University, and Huzhou Key Laboratory of Autonomous System, Huzhou 313000, China. }
        \thanks{$^3$Shanghai Institute of Special Equipment Inspection and Technical Research Co., Ltd, Shanghai 200062, China. }
        \thanks{Email:\tt\fontsize{7.8pt}{10pt}\selectfont\{mkzhang233, cxu, yanjunhi\}@zju.edu.cn}

        \vspace{-0.9cm}
}

\IEEEauthorblockN{}

\maketitle

\begin{abstract}
  Safe and efficient trajectory planning in unknown, cluttered 3D environments constitutes a critical bottleneck for deploying Unmanned Aerial Vehicles (UAVs) in real-world applications. 
  This challenge is further exacerbated by the limited field-of-view (FOV) and sensing range of onboard sensors.
  Many existing methods either make simplistic assumptions about unexplored space or rely on conservative heuristics such as speed limits or fixed perception patterns, reducing efficiency and generalizing poorly across different sensor types.
  In this work, we propose a novel planning framework that directly integrates active perception into trajectory optimization, thereby improving safety while preserving efficiency. 
  The perception constraints are derived from the UAV's dynamic model and formulated in the sensor coordinate frame, which enables precise handling of FOV geometry.
  The velocity-triggered activation mechanism enables the planner to balance perception and motion efficiency.
  We introduce an active perception sub-trajectory segment with parametric start-time optimization, mitigating collision risks from late obstacle detection.
  Our formulation enables active perception during arbitrary 3D maneuvers, extending beyond prior methods designed mainly for horizontal motion.
  All constraints and penalties are incorporated into a differentiable optimization problem, so the planner requires only a simple front-end global path for guidance, rather than a computationally expensive perception-aware path generator.
  Extensive simulations and real-world experiments demonstrate robust performance across diverse unknown environments with varying sensor configurations.
\end{abstract}

\def\abstractname{Note to Practitioners}
\begin{abstract}
The deployment of Unmanned Aerial Vehicles (UAVs) in unknown, cluttered 3D environments is constrained by limited sensor field-of-view (FOV). 
This limitation becomes particularly critical during 3D maneuvers, where limited vertical FOV and attitude changes may leave overhead or lower regions unobserved.
Existing planning methods typically involve a trade-off between safety and efficiency, often resulting in either elevated risk or reduced efficiency.
This work presents a planning framework that integrates active perception into trajectory optimization, enabling UAVs to proactively observe unknown regions before entering them.
Our method balances perception requirements and motion efficiency by optimizing an active perception segment within the trajectory, avoiding overly conservative behavior along the entire path.
Extensive simulations and real-world experiments demonstrate the framework's effectiveness across various scenarios, utilizing multiple sensor configurations such as LiDAR and vision sensors. 
This capability is particularly valuable for applications where building a prior map is difficult or impossible, such as infrastructure inspection and search-and-rescue missions.
Our planner enables complex vertical maneuvers without pre-existing maps, which is valuable in confined industrial environments and other limited-visibility structures. 
Future work will explore multi-sensor fusion and enhance the front-end component.
\end{abstract}


\section{Introduction}

Safe autonomous flight in unmapped, highly cluttered environments would greatly broaden the applicability of Unmanned Aerial Vehicles (UAVs) in missions such as exploration and search-and-rescue. 
In these settings, where UAVs should rely entirely on onboard sensing and lack any prior map, the planner must actively manage risks from unknown space. 
Therefore, it must ensure safety in the presence of both known and unknown obstacles, while maintaining high operational efficiency through active perception.

\begin{figure}[t]
  \centering
  \includegraphics[width=\linewidth]{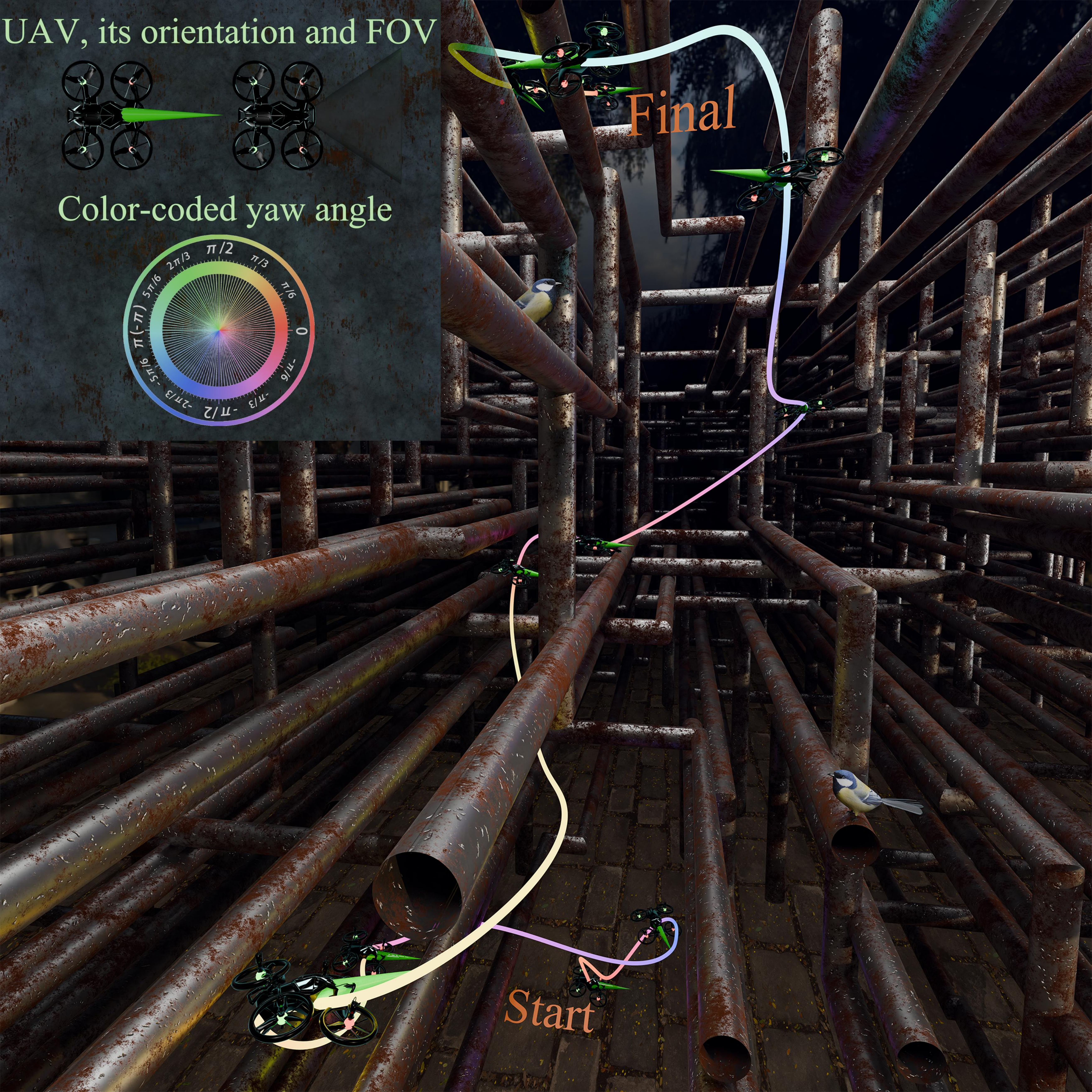}
  \caption{
    Simulation results of the proposed method in a dense grid of metal pipes without prior mapping. 
    The UAV starts from the ground, actively perceives unknown spaces using its onboard vision sensor, successfully avoids all obstacles and reaches the final position above the pipes.
  }
  \label{fig:HeadFigure}
  \vspace{-0.5cm}
\end{figure}

Existing UAV planning methods often underestimate the safety risks of unknown space.
Typically, yaw is aligned with velocity to keep the upcoming path within the FOV, yet portions of that path may still remain unobserved.  
This issue arises when the commanded turn rate exceeds the achievable yaw rate of the platform, or when high vertical speed under a limited vertical FOV causes overhead regions to remain unobserved.
Enforcing an emergency stop within the currently visible region provides a conservative safety margin, but it significantly reduces operational efficiency and, under non-omnidirectional FOV constraints, may trap the UAV in local dead ends.
Moreover, these FOV-based approaches rely on computationally expensive front-ends to generate feasible perception-aware paths, which cannot guarantee real-time performance in dense environments.
Optimization-based alternatives reduce this dependency by incorporating active perception and safety terms, such as constraining yaw/pitch to keep unknown regions in view or bounding speed.
However, these terms are often imposed only near the known--unknown boundary, which triggers observation too late and forces the UAV to slow down before entering unknown space.
Additionally, since these methods neglect the volume of the UAV, they may fail to account for unknown obstacles just outside the FOV that can still collide with the vehicle.

The main limitations are oversimplified perceptual models and overly conservative perception timing.
Existing models miss the coupling between full-attitude dynamics and sensor geometries, and they ignore how acceleration-induced pitch deflects the sensor boresight from the flight direction.
Consequently, planners must impose conservative velocity limits to prevent the UAV from rushing into unobserved regions.
Perception is typically treated as a last safety check rather than being proactively adjusted along the trajectory. 
This combination prevents planners from actively reducing environmental uncertainty, forcing an inevitable trade-off among safety, efficiency, and observability.
More critically, they lack precise modeling of the vertical FOV limits, ignoring blind regions during vertical maneuvers. 
To overcome these limitations, we propose \ACRONYM{}, a \textbf{F}ie\textbf{l}d-of-view-constrained \textbf{A}ctive \textbf{P}erception trajectory planning method for prior-map-free navigation in fully 3D environments.
\ACRONYM{} embeds active perception directly into trajectory optimization, treating it as an integral, differentiable component rather than a separate module.
This tight coupling lets \ACRONYM{} maintain both safety and navigational efficiency without resorting to conservative constraints.

We derive precise, sensor-frame perception constraints from the UAV dynamic model to ensure active observation of unknown regions along the trajectory.
We formulate active perception as risk-aware penalty terms activated by a velocity-dependent safety criterion.
As these visibility constraints become harder to satisfy, the coupled penalties naturally shift the optimized trajectory toward safer and slower motion, while allowing faster motion when observation remains feasible.
Furthermore, we insert an active perception segment into the trajectory and treat its start time as a variable to be optimized.
It encourages earlier observation of unknown regions, thereby avoiding hazards caused by late observation.
All constraints and parameters are formulated as differentiable penalty terms or unconstrained parameters, enabling straightforward integration with gradient-based optimizers.
\ACRONYM{} eliminates the dependency on a computationally heavy front-end for perception-aware path generation.
We validate \ACRONYM{} through extensive tests across diverse environments, sensor types, and mounting configurations.
Results show that \ACRONYM{} maintains safety while improving navigation efficiency in unknown environments.

Our main contributions are summarized as follows:

1. We formulate perception constraints directly in the sensor frame from the UAV dynamic model, yielding a generic FOV-aware formulation that accommodates different sensor placements and FOV geometries while supporting safety-aware planning.

2. We introduce a risk-aware active perception penalty function with a velocity-triggered activation mechanism.
Unlike conservative speed-limiting strategies, the proposed formulation allows the UAV to maintain high speed when visibility can be preserved, and leads to deceleration when maintaining visibility becomes difficult.

3. We propose a differentiable trajectory-optimization framework for arbitrary 3D maneuvers that parameterizes and optimizes the active perception segment, encouraging early observation and reducing the risk of late perception near unknown regions.

4. We validate the proposed framework in simulation and real-world experiments with limited FOV and complex 3D structures, demonstrating robust and efficient prior-map-free navigation.

The rest of the paper is organized as follows.
In Sec.~\ref{sec:Related_Work}, we review related works on trajectory planning in unknown environments.
In Sec.~\ref{sec:preliminary}, we present the preliminary concepts, including environmental representation with dual ESDF maps, front-end path planning, and the UAV dynamic model.
In Sec.~\ref{sec:activeCons}, we introduce the core active perception penalty function, detailing the visibility and safety criteria.
In Sec.~\ref{sec:Optimization}, we formulate the unified trajectory optimization problem.
Sec.~\ref{sec:simulations} and Sec.~\ref{sec:real_world} present extensive benchmark simulations and real-world experiments, respectively, to validate the proposed method.
Finally, Sec.~\ref{sec:conclusion} concludes the paper and discusses future work.

\section{Related Work}\label{sec:Related_Work}

Trajectory planning in unknown environments is fundamentally challenged by the lack of prior map knowledge.
This creates the primary risk of collision with undetected obstacles.
Accordingly, a planner must balance perception with goal-directed motion.
Safety requires sufficient time to observe the environment, which often leads to cautious, slow, or even halted motion, whereas efficiency requires rapid and direct progress toward the goal.
Here, we define active perception as planned observation of unknown regions to reduce environmental uncertainty and thereby improve trajectory safety.
Existing approaches to this challenge can be broadly categorized into three paradigms: environmental assumption, risk response, and information acquisition.

\subsection{Planners Based on Environmental Assumptions} 
A common approach is to make prior assumptions about unknown spaces, thereby converting planning into a deterministic problem.

The optimistic principle assumes all unknown spaces are free and traversable~\cite{zhou2020ego, gao2019flying, duong2020autonomous, zhang2025universal}, which encourages exploration and yields computational efficiency by avoiding complex perception and reasoning.  
However, this assumption often fails to hold in practice, thus potentially leading the vehicle into occupied regions and resulting in collisions.

Conversely, the pessimistic principle treats all unknown spaces as occupied or dangerous. 
It commonly restricts trajectories to the known free space~\cite{connolly1985determination, oleynikova2018safe, schouwenaars2002safe, bajcsy2019efficient, bes2024dwa}, decelerating to rest before entering unknown regions~\cite{liu2016high}, or penalizing paths that cross unknown spaces~\cite{zhou2020toward}.
Another approach employs motion primitives that confine motion within the previously observed FOV~\cite{lopez2017aggressive, roelofsen2017collision}. 
Although ensuring safety, such methods often induce overly conservative behavior.
In cluttered environments, the vehicle may get trapped in dead ends due to these self-imposed restrictions.

Exploration planners such as next-best-view~\cite{witting2018history} or frontier-based methods~\cite{wang2019autonomous, zhou2021fuel} alleviate these deadlocks by actively choosing viewpoints within the known space. 
Although effective in exploration, their reliance on pessimistic assumptions still compromises trajectory efficiency.

\subsection{Planners Based on Risk Response}
Instead of attempting to eliminate environmental uncertainty, risk-response planners manage navigation risk through the incorporation of a safety criterion into planning objectives or constraints.

Some methods regulate speed based on obstacle proximity, for example by imposing velocity constraints near obstacles~\cite{quan2021eva, wang2023speed} or learning maximum admissible speeds from environmental features~\cite{zhao2024learning}. 
Another strategy constrains the angle between the motion direction and the robot's orientation to mitigate risk in planar navigation~\cite{lu2021flight, lu2022real}. 
While beneficial, these heuristic rules only reduce risk, rather than directly modeling the safety of the planned trajectory.

Probabilistic methods provide a less conservative alternative by reasoning about collision likelihood under uncertainty.
Some approaches estimate collision risks in unknown spaces~\cite{richter2017bayesian} or predict unobserved obstacles from prior information~\cite{heiden2017planning, miller2021planning, katyal2021high}.
More rigorous uncertainty-aware frameworks jointly model localization, mapping, and motion uncertainty and plan in the belief space with probabilistic safety guarantees~\cite{pairet2021online}.
However, these methods primarily manage navigation risk by evaluating probabilistic collision constraints, rather than explicitly enforcing active observation of unknown regions under sensor FOV and occlusion constraints.

An alternative strategy maintains multiple trajectories with different risk levels.
For instance, some methods pair a safe path confined to the known space with an exploratory path that traverses unknown regions~\cite{tordesillas2021faster, saccani2022multitrajectory}. 
By optimizing the branching point, the planner can pursue efficiency while preserving a safe option~\cite{ren2025safety}. 
A critical limitation is that the safe trajectory can still lead to conservative dead ends if the exploratory path is blocked, especially by vertical unknown regions outside the sensor FOV that cannot be observed via yaw adjustments.

\subsection{Planners Based on Information Acquisition}

Information-acquisition planners directly address environmental uncertainty through active perception by explicitly incorporating map uncertainty reduction into the planning objective alongside safe navigation.

Some works compute information gain along trajectories, for example, by assessing how much each viewpoint reveals about occluded regions~\cite{gilhuly2024estimating} or by directly minimizing the extent of occluded spaces~\cite{higgins2021negotiating}.
Janson et al.~\cite{janson2018safe} introduce the concept of inevitable collision states (ICS), which naturally produces anticipatory behaviors such as early deceleration or detouring for observation. 
However, such approaches fail to address FOV limitations.

Another category of methods imposes perceptual constraints on the trajectory.
For instance, enforcing that the path between successive waypoints lies entirely within the FOV of the former waypoint~\cite{nieuwenhuisen2019search, yu2022cpa} prevents the robot from entering unobserved regions. 
Some methods introduce constraints that  actively steer the robot away from obstacles that cause occlusions~\cite{richter2017learning, zhou2021raptor}, thereby improving safety. 
Recent efforts combine velocity constraints with perception objectives to balance safety and information gain~\cite{yu2022cpa, higgins2022model}. However, they are limited to 2D assumptions or yaw-based horizontal sensing, failing to address vertical FOV restrictions.

Our method extends information-acquisition-based planning by embedding active perception directly into trajectory optimization, rather than treating it as a hard geometric limit or a separate information-gain objective. 
By deriving the perception constraints from the UAV dynamic model in the sensor frame, it handles general FOV geometries and supports vertical maneuvers beyond prior 2D or yaw-only methods. 
Moreover, the velocity-triggered activation mechanism and optimized active perception segment allow the planner to balance observation and deceleration adaptively, avoiding overly conservative speed limits and reducing the risks associated with late detection.

\section{Preliminary}\label{sec:preliminary}

This section presents the background concepts used in \ACRONYM{}, including environmental representation with dual ESDF maps, front-end path planning, and the UAV dynamic model.

\subsection{Dual ESDF Maps}
To obtain geometric distance information from the environment and provide differentiable constraints for trajectory optimization, we employ Euclidean Signed Distance Fields (ESDFs).
Given point clouds from sensors, we first construct a probabilistic occupancy grid map via ray-casting.
By applying a probability threshold, we partition the map into three cell types: \textit{safe}, \textit{obstacle}, and \textit{unknown}.
To address different safety constraints in planning, we build two separate ESDFs:
\begin{itemize}
    \item $E_s(\cdot)$: Treats both \textit{unknown} and \textit{obstacle} grids as unsafe.
    \item $E_u(\cdot)$: Treats only \textit{obstacle} grids as unsafe. 
\end{itemize}

\begin{figure}[t]
    \centering
    \includegraphics[width=\linewidth]{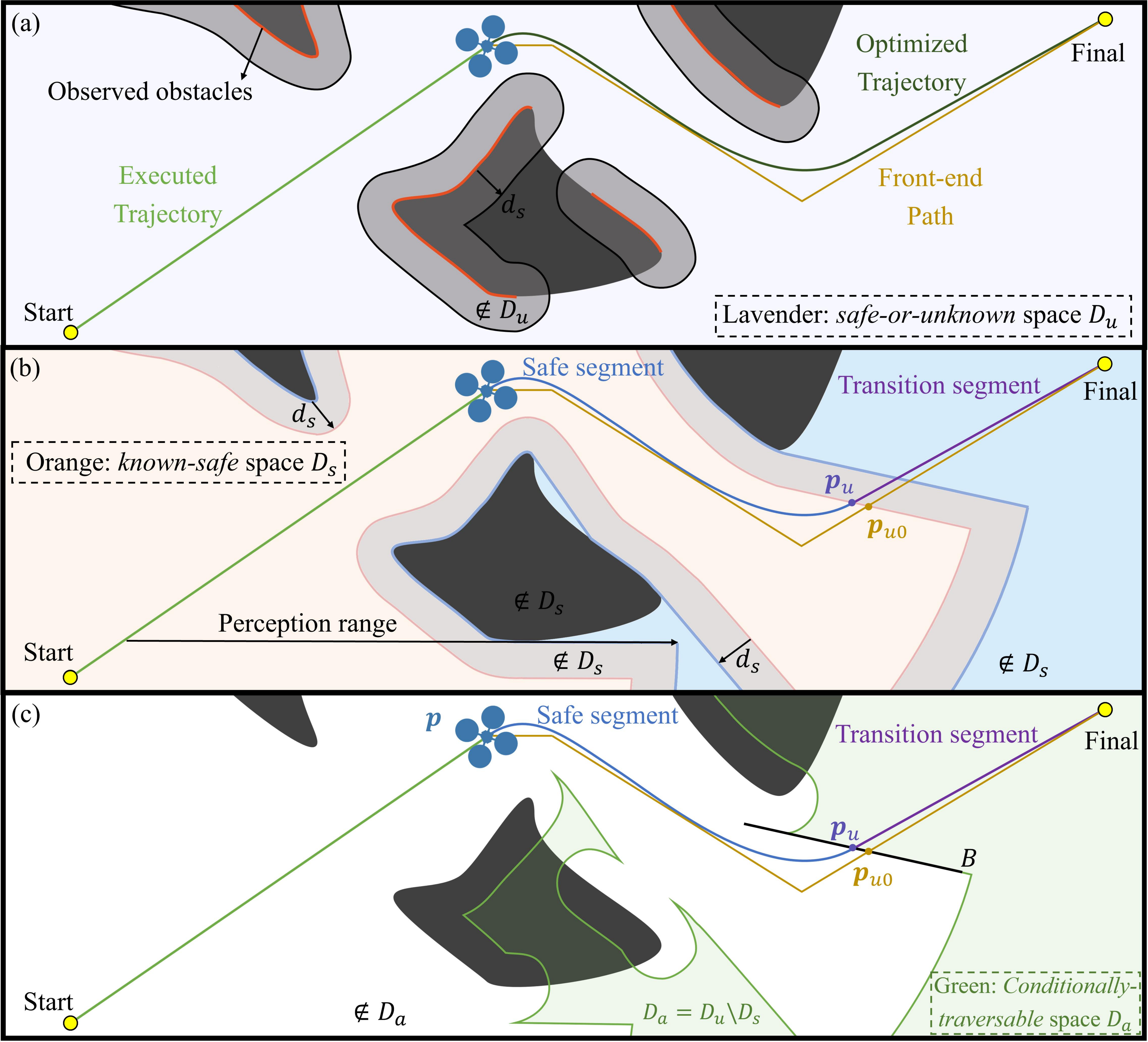}
    \caption{
      Independent visualization of the three planning spaces: (a) \textit{safe-or-unknown} space $\mathcal{D}_u$, (b) \textit{known-safe} space $\mathcal{D}_s$, (c) \textit{conditionally-traversable} space $\mathcal{D}_a$.
      We partition the trajectory into two parts: $\boldsymbol{S}_s$ in $\mathcal{D}_s$ and $\boldsymbol{S}_t$ in $\mathcal{D}_a$ or $\mathcal{D}_s$, separated by the known-unknown boundary plane $\mathcal{B}$.
      The boundary point between safe segment $\boldsymbol{S}_s$ and transition segment $\boldsymbol{S}_t$ is initialized as $\boldsymbol{p}_{u0}$ and is optimized on $\mathcal{B}$.
      }
      \vspace{-0.4cm}
    \label{fig:known_unknown}
\end{figure}

To account for the UAV's physical volume, we define a safety margin $d_s$.
As shown in Fig.~\ref{fig:known_unknown}, we define the \textit{known-safe} space $\mathcal{D}_s$ and \textit{safe-or-unknown} space $\mathcal{D}_u$ based on the ESDF values and the safety margin $d_s$:
\begin{align}
    \mathcal{D}_s &= \{\boldsymbol{p} \mid E_s(\boldsymbol{p}) \geq d_s\}, \\
    \mathcal{D}_u &= \{\boldsymbol{p} \mid E_u(\boldsymbol{p}) \geq d_s\}.
\end{align}
We further define $\mathcal{D}_a = \mathcal{D}_u \setminus \mathcal{D}_s$. 
$\mathcal{D}_a$ denotes the \textit{conditionally-traversable} space, i.e., an unknown region that is potentially safe with respect to detected obstacles but requires active perception for safe traversal. 

\subsection{Front-End Path Planning}\label{ssec:front_end}
To efficiently compute a global path that leverages the UAV's agile maneuverability, we employ a weighted A* algorithm for front-end global path planning.
Although both horizontal and vertical FOV limitations may affect perception, horizontal observability can often be improved by yaw rotation, or is naturally available for omnidirectional LiDARs. 
In contrast, vertical active perception is less efficient, because vertical observability is coupled with altitude changes and the limited vertical FOV of the sensor. 
Therefore, in the front-end search, we encourage the UAV to approach the goal height while it remains in the known-safe space, thereby reducing unnecessary entries into \textit{conditionally-traversable} space and decreasing the need for repeated vertical active perception.
We penalize front-end points that first enter \textit{conditionally-traversable} space at heights differing from the goal height $i^z_g$.
An additional penalty term $g_h$ is incorporated into the node cost $g$:
\begin{equation}
    g_h = 
    \begin{cases} 
        w_r \left| i^z_c - i^z_g \right| + w_a, & \text{if } \boldsymbol{i}_p \in \mathcal{D}_s \text{ and } \boldsymbol{i}_c \in \mathcal{D}_a, \\
        0, & \text{otherwise},
    \end{cases}
\end{equation}
where $\boldsymbol{i}_p$, $\boldsymbol{i}_c$, and $\boldsymbol{i}_g$ denote the parent, current, and goal nodes, respectively, with $i_{(\cdot)}^z$ representing the z-axis coordinate of node $\boldsymbol{i}_{(\cdot)}$.
Here, $w_r$ penalizes height deviation from the goal.
This encourages the UAV to reach near the goal height $i^z_g$ while still within known space $\mathcal{D}_s$, reducing repeated active perception.
Meanwhile, $w_a$ adds a fixed penalty for entering unknown space, prioritizing paths that exploit known-safe space before active perception becomes necessary.

\subsection{UAV Dynamic Model}
The agile maneuvers required for active perception necessitate a precise UAV dynamic model to ensure accurate constraint formulation.
Following~\cite{wang2022robust}, we define the control input as $\boldsymbol{u}=\{f, \boldsymbol{\omega}\}$, where $f\in\mathbb R_{\geq 0}$ is the collective thrust and $\boldsymbol{\omega}\in\mathbb R^3$ is the body rate. 
The dynamics are defined as:
\begin{align}
    \begin{cases} 
        m\dot{\boldsymbol{v}} = -mg\boldsymbol{e}_3 - \boldsymbol{RDR}^\top\sigma(\|\boldsymbol{v}\|_2)\boldsymbol{v} + \boldsymbol{R}f\boldsymbol{e}_3, \\
        \dot{\boldsymbol R} = \boldsymbol R\hat{\boldsymbol\omega}, 
    \end{cases}
\end{align}
where $m$ is the mass, $\boldsymbol{v}$ is velocity, $g$ is gravitational acceleration, and $\boldsymbol{e}_3 = (0,0,1)^\top$. 
$\boldsymbol{R}$ denotes the rotation matrix from the body frame to the world frame, and $\hat{\boldsymbol{\omega}} \in \mathbb{R}^{3 \times 3}$ is the skew-symmetric matrix form of the body rate $\boldsymbol{\omega} \in \mathbb{R}^3$. 
The drag coefficient matrix is given by $\boldsymbol{D} = \mathrm{diag}(d_h, d_h, d_v) \in \mathbb{R}^{3 \times 3}$, and $\sigma(\|\boldsymbol{v}\|_2)=1+c_p\sqrt{\|\boldsymbol{v}\|_2^2+\epsilon_v}$ is a nonlinear term that captures dynamic effects not accounted for by linear models, where $c_p$ is the parasitic drag coefficient and $\epsilon_v$ is a small smoothing factor.

The body-frame z-axis, which coincides with the thrust direction, is given by:
\begin{equation}
    \boldsymbol{z}_b = \mathcal{N}(\dot{\boldsymbol{v}} + g\boldsymbol{e}_3 + \frac{d_h}{m}\sigma(\|\boldsymbol{v}\|_2)\boldsymbol{v}), \quad \mathcal{N}(\boldsymbol{x}) := \boldsymbol{x} / \|\boldsymbol{x}\|_2.
\end{equation}
We decompose the attitude quaternion via Hopf fibration into a yaw component $\boldsymbol{q}_\psi$ and a tilt component $\boldsymbol{q}_z$:
\begin{align}
  \boldsymbol{q}_\psi &= (\cos(\psi/2), 0 , 0 , \sin(\psi/2))^\top \label{ali:qpsi}, \\
  \boldsymbol{q}_z &= \frac{1}{\sqrt{2(1 + \boldsymbol{z}_b^3)}} (1 + \boldsymbol{z}_b^3 , -\boldsymbol{z}_b^2 , \boldsymbol{z}_b^1 , 0)^\top,  \label{ali:qz}
\end{align}
where $z_b^i$ denotes the $i$-th component of $\boldsymbol{z}_b$.
The full attitude quaternion $\boldsymbol{q}$ is:
\begin{align}
  \boldsymbol{q} = \boldsymbol{q}_z \otimes \boldsymbol{q}_\psi, 
  \label{eq:quat_full}
\end{align}
where $\otimes$ denotes quaternion multiplication.
The corresponding rotation matrix is $\boldsymbol{R}(\boldsymbol{q})=\mathcal{R}_{\text{quat}}(\boldsymbol{q})$, where $\mathcal{R}_{\text{quat}}(\cdot)$ denotes the quaternion-to-rotation-matrix mapping.

\section{Active Perception Penalty Function}\label{sec:activeCons}
To embed active perception within the trajectory optimization framework, we formulate a set of risk-aware perception penalty functions. 
The penalty function is derived from geometric visibility criteria and safety margins, ensuring that the UAV proactively reduces environmental uncertainty while maintaining collision-free operation.

\subsection{Known-Unknown Space Boundary} \label{subsec:unknown_boundary}
After obtaining the front-end path, we identify the transition point $\boldsymbol{p}_{u0}$ as the first point at which the path enters $\mathcal{D}_a$ from $\mathcal{D}_s$, as shown in Fig.~\ref{fig:known_unknown}. 
However, the current boundary between $\mathcal{D}_s$ and $\mathcal{D}_a$ is not necessarily determined solely by the instantaneous occlusion along the line segment from the current UAV position $\boldsymbol{p}_c$ to $\boldsymbol{p}_{u0}$, because the region around $\boldsymbol{p}_{u0}$ may already have been updated by previous observations. We therefore determine the boundary from the ESDF of the latest map.

To define the boundary, we consider a cubic region spanning $I$ voxels per side centered at $\boldsymbol{p}_{u0}$. 
Within this region, we identify all grid points classified as \textit{known-safe} that are adjacent to at least one unknown neighbor which belongs to $\mathcal{D}_a$.
These points collectively define the boundary between known and unknown spaces.
Next, we fit a plane by these boundary points and calculate its normal vector $\boldsymbol{n}_u$ oriented toward the safe space.
The equation of the boundary plane is given by:
\begin{equation}
    \mathcal{B}: \boldsymbol{n}_u^\top (\boldsymbol{p}_\mathcal{B} - \boldsymbol{p}_{u0}) = 0,
\end{equation}
where $\boldsymbol{p}_\mathcal{B}$ is any point on the plane. 

As shown in Fig.~\ref{fig:known_unknown}(c), we partition the trajectory to be optimized into two contiguous parts:
\begin{itemize}
\item \textbf{Safe segment} $\boldsymbol{S}_s$: The first $m_1$ segments in known space $\mathcal{D}_s$.
\item \textbf{Transition segment} $\boldsymbol{S}_t$: The subsequent $m_2$ segments that may traverse \textit{conditionally-traversable} space $\mathcal{D}_a$.
\end{itemize}
These two segments are separated by the boundary plane $\mathcal{B}$. 
The total number of trajectory segments is $M=m_1+m_2$.
The point $\boldsymbol{p}_u$ on $\mathcal{B}$ typically represents where the trajectory transitions from known to unknown space.

\subsection{Visibility Criteria for Unknown Points} \label{subsec:visibility_criteria}

\begin{figure}[t]
    \centering
    \subfloat[
    Visualization of the viewing-angle constraint used to avoid grazing observations near the known--unknown boundary.
    Although shown as a horizontal schematic for clarity, the same formulation applies in 3D space.
    \label{fig:safe_ori_cons}]{
        \includegraphics[width=\linewidth]{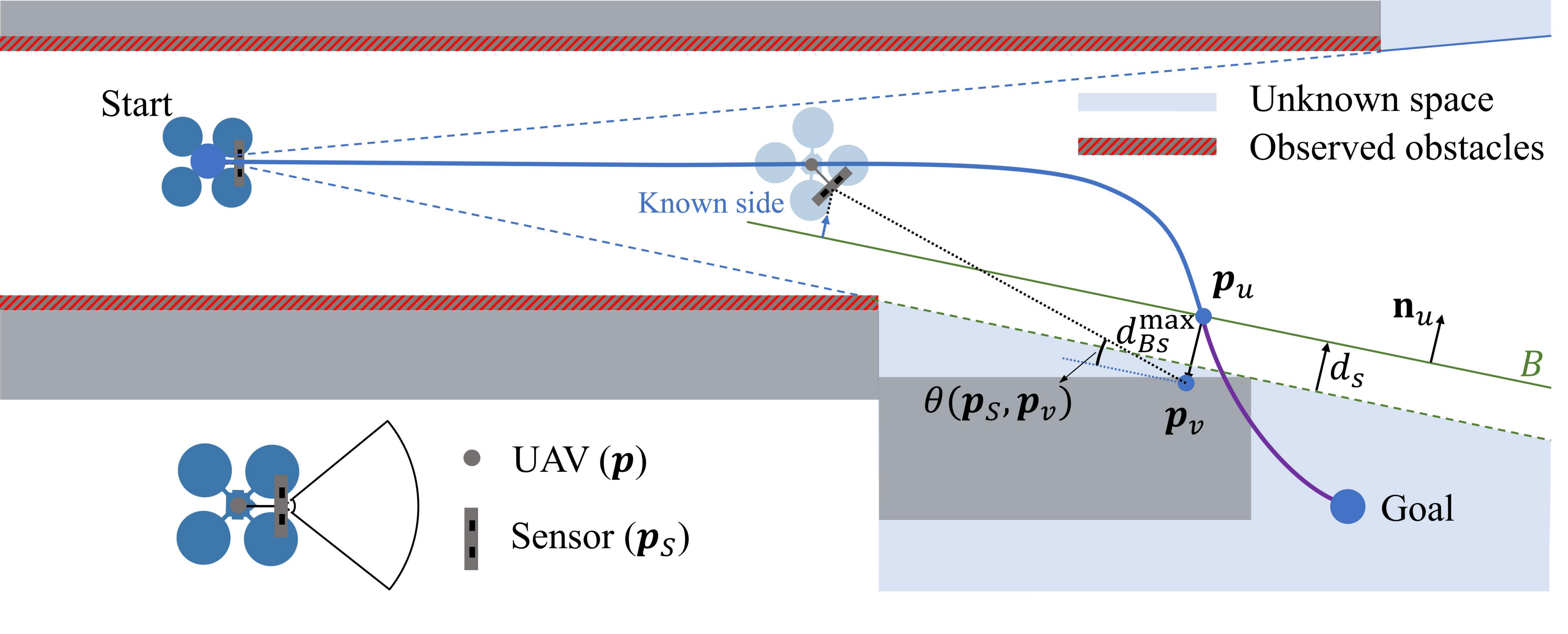}
    }\\[3pt]
    \vspace{-0.3cm}
    \subfloat[
    Visualization of the horizontal and vertical visibility constraints.
    \label{fig:ver_hri}]{
        \includegraphics[width=\linewidth]{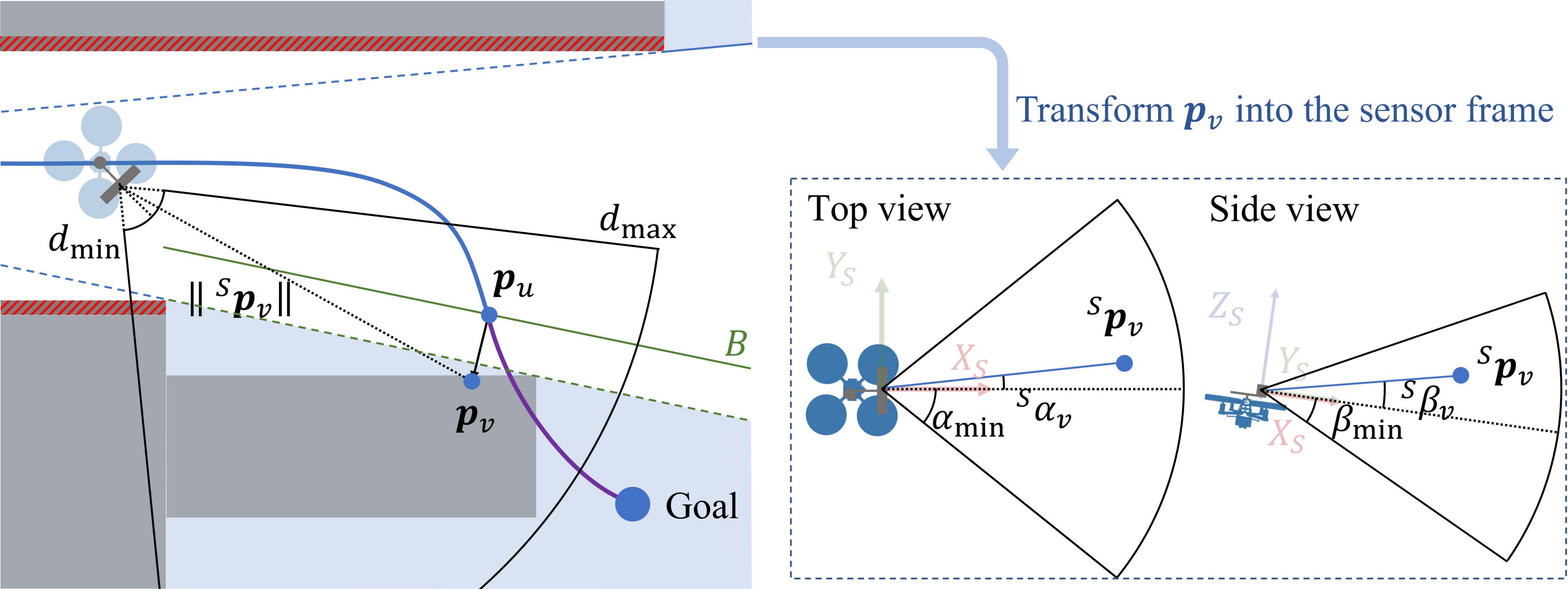}
    }
    \caption{
    Visualization of the \textit{visibility criteria} and the geometric quantities involved in their formulation.
    A forward-facing depth camera is used only for illustration, while the same sensor-frame formulation applies to other sensors by changing the FOV and range parameters.
    }
    \vspace{-0.4cm}
  \label{fig:constraints_stack}
\end{figure}

Because of the UAV's volume and the safety margin $d_s$, the boundary point $\boldsymbol{p}_u$ should not be used directly as the perception target. 
Otherwise, when the UAV reaches $\boldsymbol{p}_u$, part of its body may already extend into the adjacent unknown region, where obstacles may still be hidden, as shown in Fig.~\ref{fig:constraints_stack}(a).
We therefore define the visibility point $\boldsymbol{p}_v$ by shifting $\boldsymbol{p}_u$ along $-\boldsymbol{n}_u$, so that the safety of $\boldsymbol{p}_u$ can be evaluated while keeping the UAV safely:
\begin{equation}
    \boldsymbol{p}_v = \boldsymbol{p}_u - d_{\mathcal{B}s}^{\max} \boldsymbol{n}_u, \label{ali:pv}
\end{equation}
where $d_{\mathcal{B}s}^{\max} > d_s$ is a distance parameter that encourages the UAV to observe further into the unknown region beyond the boundary $\mathcal{B}$.

We consider a generic onboard sensor with a finite sensing range and a bounded FOV. 
Its valid sensing range is defined by a maximum distance $d_{\max}$ and a minimum distance $d_{\min}$. 
The sensor FOV is characterized by a vertical angular interval $[\alpha_{\min}, \alpha_{\max}]$ and a horizontal angular interval $[\beta_{\min}, \beta_{\max}]$, measured with respect to the sensor's principal axis.

We define the sensor frame ${S}$ as a right-handed coordinate system whose +x-axis aligns with the sensor's optical axis, whose +y-axis points laterally, and whose +z-axis points upward.
The sensor is mounted at position ${}^{B}\boldsymbol{p}_{S}$ in the body frame, and its orientation relative to the body frame is described by the rotation matrix ${}^{B}\boldsymbol{R}_{S} \in SO(3)$, which maps sensor-frame vectors to the body frame.
Let $\boldsymbol{p}$ denote the UAV position in the world frame.
For the visibility point $\boldsymbol{p}_v$ expressed in the world frame, its coordinates in the sensor frame, denoted by ${}^{S}\boldsymbol{p}_{v}=({}^{S}x_v,{}^{S}y_v,{}^{S}z_v)^\top$, are given by:
\begin{equation}
{}^{S}\boldsymbol{p}_{v}
=
({}^{B}\boldsymbol{R}_{S})^\top
\left(
\boldsymbol{R}(\boldsymbol{q})^\top(\boldsymbol{p}_v-\boldsymbol{p})
-{}^{B}\boldsymbol{p}_{S}
\right).
\end{equation}
For $\boldsymbol{p}_v$ to be visible, it must lie within the sensor's FOV, as shown in Fig.~\ref{fig:constraints_stack}(b).
This condition is expressed by the following constraints:
\begin{align}
  \mathcal{C}_{\mathrm{ver}} &= ({}^{S}\alpha_v - \frac{\alpha_{\min} + \alpha_{\max}}{2})^2 - (\frac{\alpha_{\max} - \alpha_{\min}}{2})^2 \leq 0, \notag \\
  {}^{S}\alpha_v &= \mathrm{atan2}({}^{S}z_v, \sqrt{({}^{S}x_v)^2 + ({}^{S}y_v)^2}), \label{ali:Cver} \\
  \mathcal{C}_{\mathrm{hor}} &= ({}^{S}\beta_v - \frac{\beta_{\min} + \beta_{\max}}{2})^2 - (\frac{\beta_{\max} - \beta_{\min}}{2})^2 \leq 0,  \notag\\
  {}^{S}\beta_v &= \mathrm{atan2}({}^{S}y_v, {}^{S}x_v), \label{ali:Chor}
\end{align}
where ${}^{S}\alpha_v$ and ${}^{S}\beta_v$ denote the vertical and horizontal viewing angles from the sensor's optical axis to the point ${}^{S}\boldsymbol{p}_v$, respectively.
$\mathcal{C}_\cdot$ denotes an instantaneous constraint function evaluated at a state, and the subscript specifies the particular type of constraint.

\begin{figure}[h]
    \centering
    \vspace{-0.2cm}
    \includegraphics[width=\linewidth]{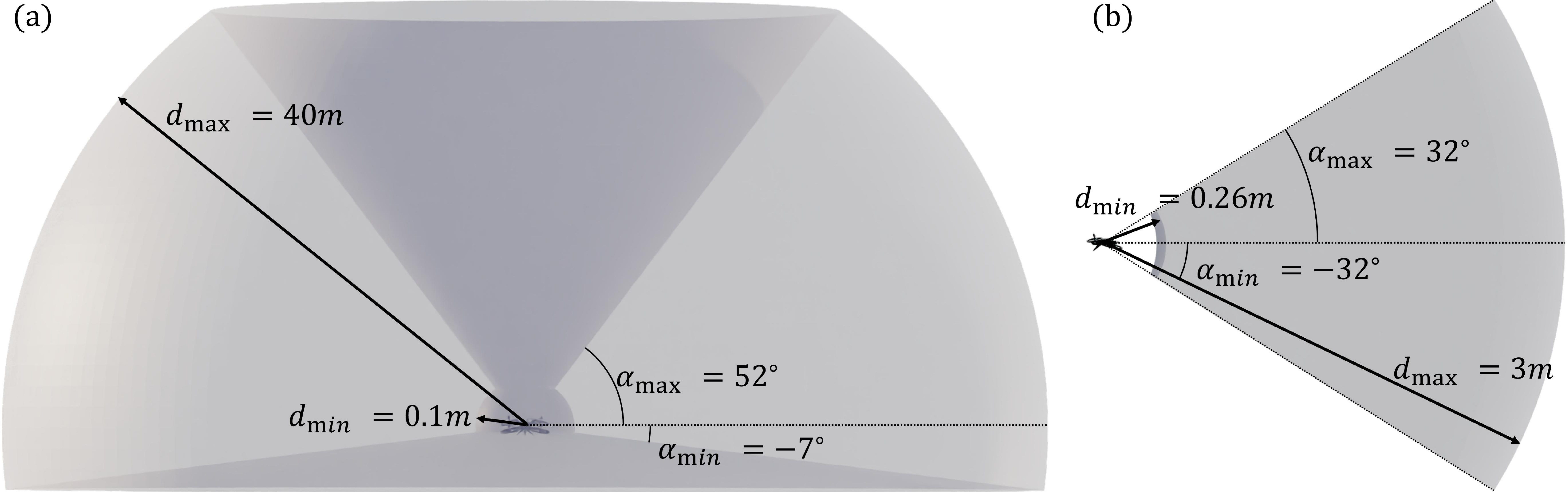}
    \caption{
      Two representative sensor configurations.
      (a) Asymmetric vertical angular coverage of the Livox Mid-360 LiDAR.
      (b) Symmetric vertical angular coverage of a typical depth camera.
      }
    \vspace{-0.2cm}
    \label{fig:radar_camera_FOV}
\end{figure}

For concreteness, consider two common sensor configurations, as shown in Fig.~\ref{fig:radar_camera_FOV}: 
\begin{itemize}
\item Depth Camera: A typical monocular or stereo depth camera (e.g., Orbbec Gemini 335\footnote{\url{https://www.orbbec.com/products/stereo-vision-camera/gemini-335/}}) features a horizontal FOV of approximately $78^\circ$ and a vertical FOV of approximately $64^\circ$, corresponding to $\alpha_{\max} = -\alpha_{\min} = 32^\circ$ and $\beta_{\max} = -\beta_{\min} = 39^\circ$, with a recommended sensing range of $0.26$ to $3$ m. 
\item 3D LiDAR: A Livox Mid-360 LiDAR\footnote{\url{https://www.livoxtech.com/mid-360}}  features a horizontal FOV of $360^\circ$ and a non-symmetric vertical FOV spanning approximately $52^\circ$ upward and $7^\circ$ downward, with a typical sensing range of $0.1$ to $40$ m. 
In our formulation, this is modeled as $\alpha_{\min} = -7^\circ$, $\alpha_{\max} = 52^\circ$ with the horizontal FOV constraint disabled.
\end{itemize}

These examples demonstrate the flexibility of our generic sensor model to accommodate diverse hardware configurations without modification to the core constraint formulation.

Considering the limitation of the sensing range, we constrain the distance from ${}^{S}\boldsymbol{p}_v$ to the origin of the sensor coordinate system:
\begin{equation}
    \begin{aligned}[c]
  \mathcal{C}_{\mathrm{d}_{\max}} = \|{}^{S}\boldsymbol{p}_v\|_2 - d_{\max} \leq 0,\\
  \mathcal{C}_{\mathrm{d}_{\min}} = d_{\min} - \|{}^{S}\boldsymbol{p}_v\|_2 \leq 0. 
    \end{aligned}
\end{equation}
Since unknown spaces may be occluded by obstacles adjacent to the boundary, we encourage the sensor to remain on the safe side of $\mathcal{B}$ to observe $\boldsymbol{p}_v$: 
\begin{align}
  & \boldsymbol{p}_{S} = \boldsymbol{p} + \boldsymbol{R}(\boldsymbol{q}) {}^{B}\boldsymbol{p}_{S}, \\
  & \mathcal{C}_{\mathrm{vd}} = -\boldsymbol{n}_u^\top (\boldsymbol{p}_{S} - \boldsymbol{p}_u) \leq 0, \label{ali:Cvd}
\end{align}
where $\boldsymbol{p}_{S}$ is the sensor position in the world frame.

Additionally, observing $\boldsymbol{p}_v$ at an overly grazing angle degrades perception quality. 
This is primarily because a fixed sensor angular resolution projects onto a larger surface area as the incidence angle increases, reducing spatial resolution. 
Consequently, depth measurements become noisier and less accurate. 
To prevent this, we ensure the sensor maintains sufficient distance from the plane parallel to $\mathcal{B}$ passing through $\boldsymbol{p}_v$, as shown in Fig.~\ref{fig:constraints_stack}(a).
To ensure effective perception, we introduce a minimum viewing angle $\theta_v$: 
\begin{align}
  \theta(\boldsymbol{p}_{S}, \boldsymbol{p}_v)  = \arcsin\left( \frac{\mathrm{dis}(\mathcal{B}, \boldsymbol{p}_{S}) + d_{\mathcal{B}s}^{\max}}{\|\boldsymbol{p}_{S} - \boldsymbol{p}_v\|_2} \right) \geq \theta_v,
\end{align}
where $\theta(\boldsymbol{p}_{S}, \boldsymbol{p}_v)$ is the viewing angle between the line-of-sight from the sensor to $\boldsymbol{p}_v$ and the boundary plane $\mathcal{B}$, $\mathrm{dis}(\mathcal{B}, \boldsymbol{p}_{S}) \triangleq \boldsymbol{n}_u^\top(\boldsymbol{p}_{S}-\boldsymbol{p}_u)$ is the signed distance from the sensor position to the boundary plane along the safe-side normal, $\|\boldsymbol{p}_{S} - \boldsymbol{p}_v\|_2$ is the point-to-point distance, and $\theta_v$ is the desired minimum viewing angle.
The original constraint contains both a division by $\|\boldsymbol{p}_{S}-\boldsymbol{p}_v\|_2$ and an $\arcsin(\cdot)$ operation, which can be poorly conditioned near the feasibility boundary.
We therefore impose the following form of the same angular threshold for $\theta_v\in(0,\pi/2)$:
\begin{align}
      \mathcal{C}_{\mathrm{v}\theta} &= \|\boldsymbol{p}_{S} - \boldsymbol{p}_v\|_2 \sin\theta_v - (\mathrm{dis}(\mathcal{B}, \boldsymbol{p}_{S}) + d_{\mathcal{B}s}^{\max}) \leq 0.  \label{ali:Cvtheta}
\end{align}

The constraints formulated above form the general \textit{visibility criteria}. 
These criteria can be readily adapted to various sensor types by modifying the specific constraint formulations. 

\subsection{Risk-Aware Active Perception} \label{subsec:risk_aware_active_perception}
To ensure safety, the target point $\boldsymbol{p}_v$ must be observed before the UAV enters the unknown space at $\boldsymbol{p}_u$. 
The UAV must therefore maintain emergency braking capability, which is formalized as the \textit{safety criterion}:
\begin{equation}
    \mathcal{C}_{\mathrm{sd}} = \frac{\|\boldsymbol{v}\|_2^2}{2 a_{\max}} + d_m - \|\boldsymbol{p} - \boldsymbol{p}_u\|_2 \leq 0, \label{ali:Csd}
\end{equation}
where $\boldsymbol{v}$ is the current velocity, $d_m$ is a safety margin for replanning, and $a_{\max}$ is a conservatively chosen deceleration magnitude. 
Since the UAV's achievable maximum deceleration may vary with its operating condition, $a_{\max}$ is set below the physical limit to ensure that the planned braking maneuver remains feasible during execution.

\begin{figure}[t]
    \centering
    \includegraphics[width=\linewidth]{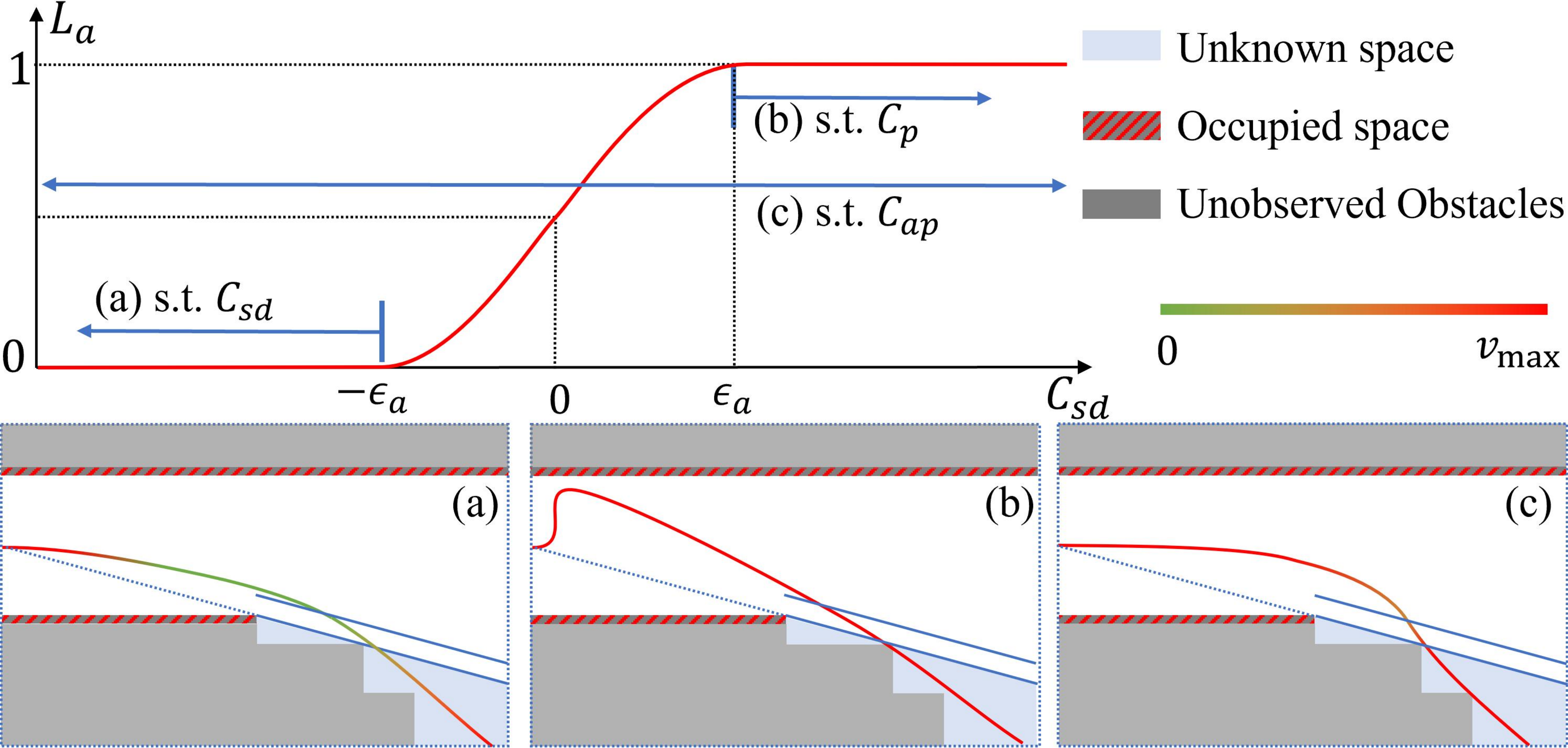}
    \caption{
      Activation function $\mathcal{L}_a(\mathcal{C}_{\mathrm{sd}})$ and the resulting trajectory behaviors in a simplified 2D illustration, where the effect of orientation is ignored and the constraints are applied over the entire $\boldsymbol{S}_s$ segment. 
      (a) Safety-dominated case with $\mathcal{L}_a(\mathcal{C}_{\mathrm{sd}})=0$. 
      (b) Perception-dominated case with $\mathcal{L}_a(\mathcal{C}_{\mathrm{sd}})=1$. 
      (c) Proposed risk-aware active perception behavior.
      }
    \vspace{-0.4cm}
    \label{fig:La}
\end{figure}

To achieve an adaptive strategy that balances safety and perception, we introduce an activation function $\mathcal{L}_a(x)$, as shown in Fig.~\ref{fig:La}:
\begin{equation}
    \mathcal{L}_a(x) = 
    \begin{cases} 
        0, & x \leq -\epsilon_a, \\
        \frac{1}{2\epsilon_a^4} (x + \epsilon_a)^3 (\epsilon_a - x), & -\epsilon_a < x \leq 0, \\
        \frac{1}{2\epsilon_a^4} (x - \epsilon_a)^3 (\epsilon_a + x) + 1, & 0 < x \leq \epsilon_a, \\
        1, & x \geq \epsilon_a,
    \end{cases}
    \label{ali:La}
\end{equation}
where $\epsilon_a \in \mathbb{R}^+$. 
This activation function acts as a continuous gate that modulates the visibility penalties according to the \textit{safety criterion}.
The active perception penalty function $\mathcal{P}_{\mathrm{ap}}$ is thus given by:
\begin{equation}
    \mathcal{P}_{\mathrm{ap}} = \mathcal{L}_a(\mathcal{C}_{\mathrm{sd}}) \cdot \sum_{\ell\in \mathbb{P}} w_{\ell} \mathcal{L}_1(\mathcal{C}_{\ell}), \label{ali:Cap}
\end{equation}
where $\mathbb{P} = \{\mathrm{ver}, \mathrm{hor}, \mathrm{d_{\min}}, \mathrm{d_{\max}}, \mathrm{vd}, \mathrm{v}\theta\}$ is the index set of the visibility constraints, $w_{\ell}$ are their respective weights, and $\mathcal{L}_1(\cdot)$ is a first-order smooth relaxation function, whose exact form follows~\cite{wang2022robust}.

As shown in Fig.~\ref{fig:La}(a), when the safety criterion is well satisfied, $\mathcal{L}_a(\mathcal{C}_{\mathrm{sd}})=0$, and the perception constraints are not enforced.
The trajectory is mainly affected by the safety-related terms, leading to conservative motion before entering the unknown space.
In Fig.~\ref{fig:La}(b), the trajectory satisfies the perception constraints for observing $\boldsymbol{p}_v$.
Since the safety-related restriction is not considered, the UAV can maintain a higher speed than in the safety-dominated case while detouring to keep $\boldsymbol{p}_v$ observable.
However, relying only on safety-related terms or only on perception requirements is generally undesirable. 
Our formulation in Eq.~\eqref{ali:Cap} uses $\mathcal{L}_a(\mathcal{C}_{\mathrm{sd}})$ to continuously weight the perception penalties according to the current collision risk.
When the safety criterion is well satisfied, the perception penalties are suppressed, allowing the UAV to continue moving safely and attempt to observe $\boldsymbol{p}_v$ from a more suitable state.
As the collision risk increases, the perception penalties are gradually activated, encouraging the UAV to observe $\boldsymbol{p}_v$.
As shown in Fig.~\ref{fig:La}(c), this mechanism enables the trajectory to adapt between safety-oriented and perception-oriented behaviors.

\section{Optimization Problem Formulation}\label{sec:Optimization}
Building upon the active perception penalty function and dynamic models, we formulate the planning task as a unified, differentiable optimization problem.
This formulation minimizes control effort while modeling safety and dynamic constraints, and integrating the active perception process.

\subsection{Trajectory Representation}
We employ the MINCO trajectory class~\cite{wang2022geometrically} to achieve minimum control effort spatiotemporal trajectory optimization.
The trajectory is represented as a temporal polynomial function $\boldsymbol{o}_i(t) \in \mathbb{R}^4$ of the differentially flat output $\boldsymbol{o}:=[\boldsymbol{p}^\top, \psi]^\top$, where $\boldsymbol{p}\in\mathbb R^3$ is the position and $\psi$ is the yaw angle. 
MINCO utilizes an invertible banded matrix $\boldsymbol{K}(\boldsymbol{T}) \in \mathbb{R}^{2Mh \times 2Mh}$ to decouple the spatial and temporal parameters of an $M$-segment trajectory $\boldsymbol{o}(t)$, where $\boldsymbol{T} = [T_1, \ldots, T_M]^\top \in \mathbb{R}_{>0}^M$ denotes the duration of each segment. 
This enables efficient spatiotemporal deformation with linear complexity in the flat output space:
\begin{align}
& \boldsymbol{K}(\boldsymbol{T})\boldsymbol{c} = [\boldsymbol{o}_0^{[h-1]}, \boldsymbol{o}_1', \boldsymbol{0}, \ldots, \boldsymbol{o}_{M-1}', \boldsymbol{0}, \boldsymbol{o}_f^{[h-1]}]^\top, \label{ali:mincoK} \\
& \boldsymbol{o}_i(t) = \boldsymbol{c}_i^\top\boldsymbol{\beta}(t), \quad \forall t \in [0, T_i],
\end{align}
where $h$ is the system order, and each segment of the trajectory is parameterized as a polynomial of degree $2h-1$. 
The matrix $\boldsymbol{o}' = [\boldsymbol{o}_1', \ldots, \boldsymbol{o}_{M-1}'] \in \mathbb{R}^{4\times(M-1)}$ collects the intermediate waypoints as columns. 
The matrix $\boldsymbol{c} \in \mathbb{R}^{2Mh \times 4}$ contains the polynomial coefficients, and $\boldsymbol{c}_i \in \mathbb{R}^{2h \times 4}$ denotes the coefficient block corresponding to the $i$-th segment.
The initial and final states are constrained by $\boldsymbol{o}_0^{[h-1]} = [\boldsymbol{o}_0, \dot{\boldsymbol{o}}_0, \ldots, \boldsymbol{o}_0^{(h-1)}]$ and $\boldsymbol{o}_f^{[h-1]}$, respectively. 
$\boldsymbol{\beta}(t) = [1, t, \ldots, t^{2h-1}]^\top$ is the natural polynomial basis.
Eq.~\eqref{ali:mincoK} enforces continuity of $\boldsymbol{o}$ and its derivatives up to order $2h-2$ across adjacent segments, and enforces the initial state constraint $\boldsymbol{o}_1^{[h-1]}(0) = \boldsymbol{o}_0^{[h-1]}$ and the final state constraint $\boldsymbol{o}_M^{[h-1]}(T_M) = \boldsymbol{o}_f^{[h-1]}$. 
In this work, we set the system order to $h=3$.

\subsection{Trajectory Truncation and Final State Relaxation}\label{ssec:truncation_and_relaxation}

Due to the need for frequent replanning during operation, trajectories are truncated to prevent excessive length, which would lead to unnecessary computations.
We set a truncation length threshold $l_u$ for the transition segment $\boldsymbol{S}_t$ in \textit{conditionally-traversable} space $\mathcal{D}_a$.
When the front-end path enters $\mathcal D_a$ and the cumulative length therein exceeds $l_u$, the path is truncated at that point, and the remainder is discarded.
The remaining parts in $\mathcal{D}_{a}$ constitute $\boldsymbol{S}_t$, and the truncation point is set as the temporary goal position $\boldsymbol{p}_{tf}$.
This retention is crucial because the temporary goal point in $\mathcal{D}_a$ forces the trajectory to extend into the unknown space. 
We observe that truncating at the boundary induces prolonged oscillations near $\mathcal{B}$, since the planner does not account for the efficiency of subsequent segments in $\mathcal{D}_a$.
This effect is particularly pronounced during vertical motion.

Following truncation, the UAV is directed toward a temporary terminal state $\boldsymbol{o}_f = \boldsymbol{o}_{tf}$ instead of the mission goal state $\boldsymbol{o}_{g}$.
Accordingly, the UAV is not required to decelerate to a full stop before reaching this temporary terminal state, and no terminal yaw constraint is imposed, since such requirements often lead to conservative motion.
Unlike conventional trajectory optimization formulations with fixed terminal conditions, we relax these terminal-state components by treating them as optimization variables rather than prescribed boundary conditions.
Specifically, the terminal yaw angle $\psi_f$ and terminal velocity $\dot{\boldsymbol{p}}_f$, which are contained in the temporary terminal state $\boldsymbol{o}_f^{[h-1]}$, are optimized rather than fixed.
Once the gradient of the objective function $\mathcal{J}$ with respect to the coefficient matrix $\boldsymbol{c} \in \mathbb{R}^{2Mh \times 4}$ has been computed, the gradients with respect to $\psi_f$ and $\dot{\boldsymbol{p}}_f$ are given by:
\begin{align}
\boldsymbol{G}
&= \boldsymbol{K}^{-T}\frac{\partial \mathcal{J}}{\partial \boldsymbol{c}}
\in\mathbb{R}^{2Mh\times 4}, \nonumber \\
\frac{\partial \mathcal{J}}{\partial \psi_f}
&= (\boldsymbol{e}_4^4)^\top \boldsymbol{G}^\top 
\boldsymbol{e}_{2Mh}^{(2M-1)h+1}, \\
\frac{\partial \mathcal{J}}{\partial \dot{\boldsymbol{p}}_f}
&= [\boldsymbol{e}_4^1,\boldsymbol{e}_4^2,\boldsymbol{e}_4^3]^\top
\boldsymbol{G}^\top \boldsymbol{e}_{2Mh}^{(2M-1)h+2},
\end{align}
where $\boldsymbol{e}_n^i$ denotes the $i$-th column of the identity matrix $\boldsymbol{I}_n$.
Here, the row indices $(2M-1)h+1$ and $(2M-1)h+2$ correspond to the terminal flat output and its first derivative, respectively.
The derivations of these gradients are provided in~\cite{wang2022geometrically}.

\subsection{Active Perception Segment}\label{subsec:active_perception_segment}

\begin{figure}[t]
    \centering
    \includegraphics[width=\linewidth]{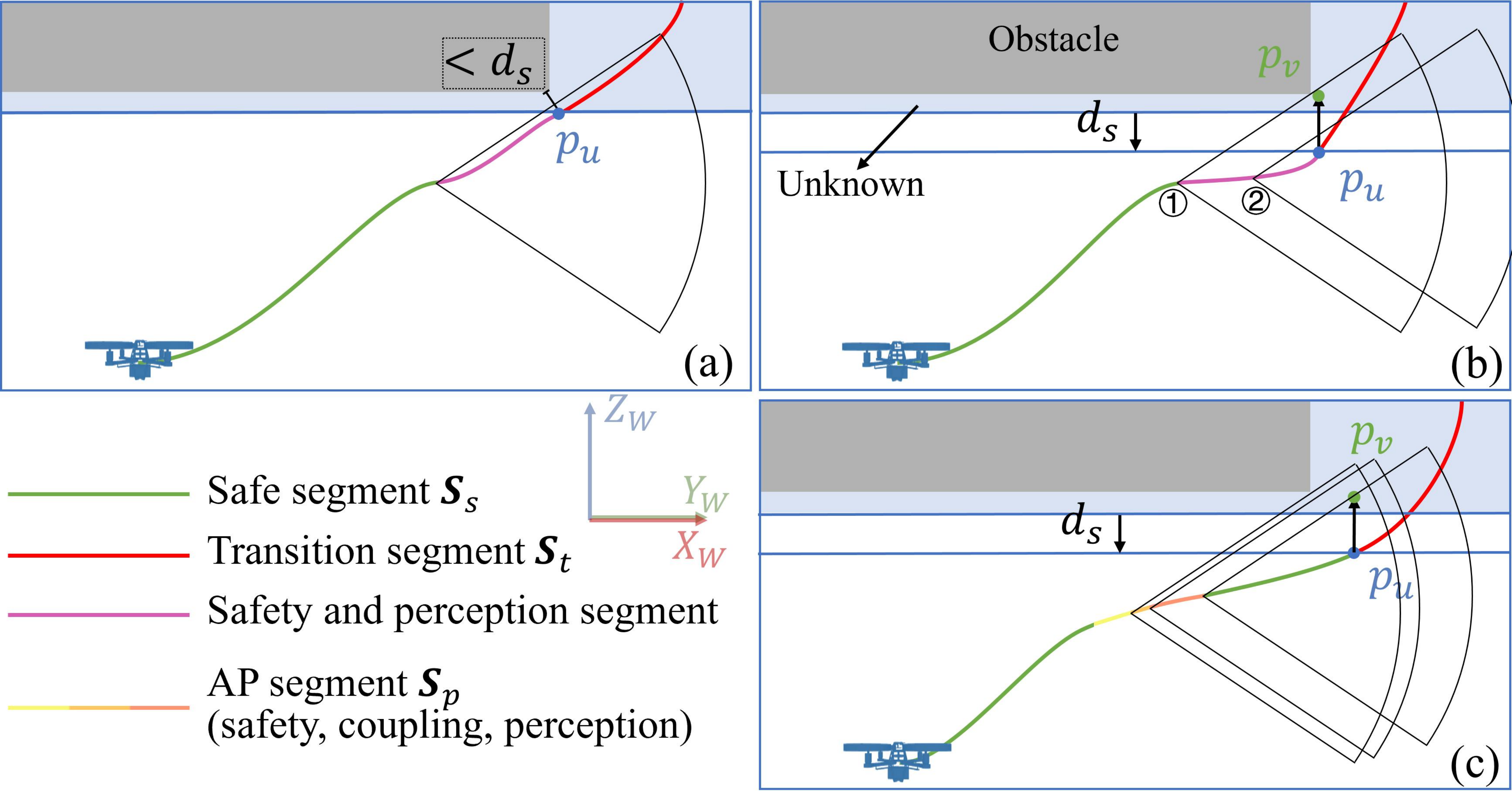}
    \caption{
        Comparison of the proposed AP segment with other strategies.
        (a) Ignoring the safety distance and evaluating the safety and perception constraints over the terminal part of $\boldsymbol{S}_s$.
        (b) Considering the safety distance and evaluating the safety and perception constraints over the terminal part of $\boldsymbol{S}_s$.
        (c) The proposed AP segment, which can be interpreted as containing safety-dominated, coupling, and perception-dominated parts induced by the risk-aware activation.
      }
    \label{fig:apsegment}
    \vspace{-0.4cm}
\end{figure}

In contrast to dynamic constraints, the active perception penalty function $\mathcal{P}_{\mathrm{ap}}$ discussed in Sec.~\ref{subsec:risk_aware_active_perception} is not suitable for the entire $\boldsymbol{S}_s$ segment. 
Some methods~\cite{yu2022cpa} apply $\mathcal{C}_{\ell}$ and $\mathcal{C}_{\mathrm{sd}}$ over the terminal part of $\boldsymbol{S}_s$.
As shown in Fig.~\ref{fig:apsegment}(a), this enforced part is marked as the safety and perception segment.
However, when the safety distance is respected, the limited FOV may cause some segments to violate the visibility constraints. 
As shown in Fig.~\ref{fig:apsegment}(b), although the UAV observes $\boldsymbol{p}_v$ at \ding{172}, the later state at \ding{173} violates the vertical visibility constraint.
Due to the limited vertical FOV and the safety margin induced by the UAV's physical volume, the UAV cannot simply move closer to the boundary to recover visibility.
Consequently, $\boldsymbol{p}_v$ lies above the sensor's observable region at \ding{173}, leaving the possible obstacle near the boundary unobserved.

Due to the limited FOV, $\boldsymbol{p}_v$ may not be observable when the UAV reaches $\boldsymbol{p}_u$, so imposing active perception at this state may introduce an active perception penalty $\mathcal{P}_{\mathrm{ap}}$ that cannot be driven to zero.
Furthermore, observing $\boldsymbol{p}_v$ requires only a short duration.
As shown in Fig.~\ref{fig:apsegment}(c), we define a sub-trajectory segment $\boldsymbol{S}_p \subset \boldsymbol{S}_s$ with a fixed duration $T_p$, called the active perception (AP) segment, over which the active perception penalty $\mathcal{P}_{\mathrm{ap}}$ is imposed.
  The start time $t_{as}$ of $\boldsymbol{S}_p$ is parameterized by $\rho \in (0,1)$, as follows:
\begin{equation}
t_{as} = \rho (T_s - T_p), \quad T_s = \sum_{i=1}^{m_1} T_i, 
\end{equation}
where $T_s$ is the duration of $\boldsymbol{S}_s$.
Next, we transform $\rho \in (0,1)$ into an unconstrained variable $\varrho \in \mathbb{R}$ as follows:
\begin{equation}
\varrho(\rho) = \log \left( \frac{\rho}{1 - \rho} \right), \qquad \rho(\varrho)=\frac{1}{1+e^{-\varrho}}.
\end{equation}
This transformation allows the optimizer to select the start time of $\boldsymbol{S}_p$. 
To encourage earlier observation, we add \( w_\rho \rho(\varrho) \) into the objective function \(\mathcal{J}\), where \( w_\rho > 0 \) is an adjustable weight.

The active perception penalty is imposed over $\boldsymbol{S}_p$ and numerically integrated by dividing the segment into $\kappa$ subintervals:
\begin{equation}
\mathcal{I}_{\mathrm{ap}} =
\sum_{j=0}^{\kappa}
\frac{T_p}{\kappa}\nu_j \,
\mathcal{P}_{\mathrm{ap}}(t_j),
\quad
t_j = t_{as} + j\frac{T_p}{\kappa}, 
\label{ali:Iap}
\end{equation}
where $(\nu_0, \nu_1, \ldots, \nu_{\kappa-1}, \nu_{\kappa}) = (1/2, 1, \ldots, 1, 1/2)$. 
Here, $\mathcal{P}_{\mathrm{ap}}(t_j)$ denotes the instantaneous active perception penalty evaluated at the sampled trajectory state at time $t_j$, i.e., the time-parameterized form of the penalty defined in Eq.~\eqref{ali:Cap}.
As shown in Fig.~\ref{fig:apsegment}(c), the AP segment $\boldsymbol{S}_p$ is optimized in its start time.
It can be interpreted as consisting of a safety-dominated phase, a coupling phase, and a perception-dominated phase, which result from the risk-aware weighting induced by $\mathcal{L}_a(\mathcal{C}_{\mathrm{sd}})$ rather than from a manually predefined partition.

Since the visibility constraints only model the FOV geometry and do not account for occlusion, $\boldsymbol{p}_v$ may remain unobserved during $\boldsymbol{S}_p$ even when these constraints are satisfied. 
In such cases, the planner may neither complete the intended observation nor trigger sufficient braking through the safety term. 
We therefore introduce a safeguard. 
Let $\boldsymbol{o}_{pf}$ denote the endpoint state of the AP segment $\boldsymbol{S}_p$.
This safeguard adds the following point-wise penalty at $\boldsymbol{o}_{pf}$, consisting of the safety term $\mathcal{C}_{\mathrm{sd}}$ and the visibility terms $\mathcal{C}_{\ell}$ for $\ell\in\mathbb{P}$:
\begin{equation}
\mathcal{S}_{\mathrm{ap}}(\boldsymbol{o}_{pf}) = w_{sd}\mathcal{L}_1(\mathcal{C}_{\mathrm{sd}}(\boldsymbol{o}_{pf})) + \sum_{\ell\in \mathbb{P}}w_{\ell}\mathcal{L}_1(\mathcal{C}_{\ell}(\boldsymbol{o}_{pf})), \label{ali:Cap_end}
\end{equation}
where $w_{sd}$ is the weight of the safety term, $\mathcal{S}(\cdot)$ denotes a point-wise penalty term, and the index set $\mathbb{P}$ is defined as in Eq.~\eqref{ali:Cap}.

\subsection{Constraints and Penalty Terms}\label{ssec:constraints}
In addition to the AP segment introduced in Sec.~\ref{subsec:active_perception_segment}, the trajectory optimization also involves a set of constraints. 
Unlike the active perception penalty function, which is imposed only over the AP segment $\boldsymbol{S}_p$, the constraints considered in this subsection are imposed mainly on the trajectory segments $\boldsymbol{S}_s$ and $\boldsymbol{S}_t$. 
Their roles are to enforce the key requirements of trajectory generation and execution, including safety, dynamic feasibility, and temporal consistency.

\subsubsection{Reparameterization with Unconstrained Variables}

These constraints are incorporated into the optimization through reparameterization, so that the optimizer can optimize unconstrained variables while the original constraints are satisfied.

\paragraph{Boundary Point Reparameterization}
As defined in Sec.~\ref{subsec:unknown_boundary}, the boundary point $\boldsymbol{p}_u$ between $\boldsymbol{S}_s$ and $\boldsymbol{S}_t$ must lie on the boundary $\mathcal{B}$.
Otherwise, the trajectory may result in observation of incorrect locations.
However, strictly satisfying this equality constraint is difficult for polynomial trajectories.
To simplify, we set $\boldsymbol{p}_u$ as the position $\boldsymbol{p}_{m_1}'$ of the intermediate waypoint $\boldsymbol{o}_{m_1}'$, which is the endpoint of $\boldsymbol{S}_s$.
Instead of constraining $\boldsymbol{p}_u$ to $\mathcal{B}$ using penalty methods, we reparameterize it by two orthogonal basis vectors lying in $\mathcal{B}$:
\begin{equation}
\boldsymbol{p}_u = \boldsymbol{p}_{\mathcal{B}} + \boldsymbol{\upsilon}_\mathcal{B}^1 \zeta^1 + \boldsymbol{\upsilon}_\mathcal{B}^2 \zeta^2, \quad \boldsymbol{p}_{\mathcal{B}} \in \mathcal{B}, \quad \zeta^1, \zeta^2 \in \mathbb{R}, \label{ali:boundary_reparameterization}
\end{equation}
where $\boldsymbol{\upsilon}_\mathcal{B}^1$ and $\boldsymbol{\upsilon}_\mathcal{B}^2$ are orthogonal basis vectors in $\mathcal{B}$ and $\boldsymbol{p}_\mathcal{B}$ is any point on the boundary plane $\mathcal{B}$.
In practice, we typically choose $\boldsymbol{p}_\mathcal{B}$ as $\boldsymbol{p}_{u0}$.
Thus, we can optimize the two unconstrained scalars $\boldsymbol{\zeta} = [\zeta^1, \zeta^2]^\top$ instead of directly optimizing $\boldsymbol{p}_u$.

\paragraph{Duration Reparameterization}
To prevent invalid durations (e.g., $T_s < T_p$ for $\boldsymbol{S}_s$, and non-positive durations for $\boldsymbol{S}_t$), we map each duration $T_i \in \mathbb{R}^{+}$ to an unconstrained variable $\tau_i$ through the following smooth bijection:
\begin{equation}
\tau_i =
\begin{cases}
\sqrt{2(T_i - T_i^{\min}) - 1} - 1, \;\; T_i - T_i^{\min} > 1, \\[4pt]
1 - \sqrt{\dfrac{2}{T_i - T_i^{\min}} - 1}, \;\; 0 < T_i - T_i^{\min} \leq 1.
\end{cases}
\label{ali:duration_mapping}
\end{equation}
With Eq.~\eqref{ali:duration_mapping}, the lower-bound constraint $T_i > T_i^{\min}$ is guaranteed throughout the optimization.
Accordingly, $\tau_i$ is optimized as an unconstrained variable, while $T_i$ is obtained from the inverse mapping of Eq.~\eqref{ali:duration_mapping}.

For $\boldsymbol{S}_s$, we set $T_i^{\min} = T_p / m_1$ to ensure that the total duration of $\boldsymbol{S}_s$ is no smaller than the active perception duration $T_p$, thereby providing sufficient time for the AP segment.
For $\boldsymbol{S}_t$, we set $T_i^{\min} = 0$.

\subsubsection{Point-wise Penalty Terms}

Some constraints are imposed directly at discrete states and are incorporated into the objective function as point-wise penalty terms, denoted by $\mathcal{S}(\cdot)$.

\paragraph{Boundary Consistency Penalties}
\label{subsec:boundary_constraints}

Since $\mathcal{B}$ is only a local planar approximation of the boundary, not every position on $\mathcal{B}$ can be regarded as a valid boundary point. 
Although $\boldsymbol{p}_u$ is constrained to lie on $\mathcal{B}$ by Eq.~\eqref{ali:boundary_reparameterization}, it may still drift beyond the locally valid boundary represented by this plane. 
Therefore, we further impose ESDF-based point-wise penalty terms to restrict $\boldsymbol{p}_u$ to the portion of $\mathcal{B}$ that remains consistent with the local boundary estimate:
\begin{equation}
\begin{aligned}[c]
\mathcal{S}_{\mathcal{B}s}^{\min}(\boldsymbol{p}_u) &= w_{\mathcal{B}s}^{\min}\mathcal{L}_1(d_{\mathcal{B}s}^{\min} - E_s(\boldsymbol{p}_u)), \\
\mathcal{S}_{\mathcal{B}s}^{\max}(\boldsymbol{p}_u) &= w_{\mathcal{B}s}^{\max}\mathcal{L}_1(E_s(\boldsymbol{p}_u) - d_{\mathcal{B}s}^{\max}),
\label{ali:boundary_esdf}
\end{aligned}
\end{equation}
where $d_{\mathcal{B}s}^{\max}$ is set to the same value as the distance used in Eq.~\eqref{ali:pv} to ensure the validity of $\boldsymbol{p}_v$ observations, and $d_{\mathcal{B}s}^{\min}$ is set to $d_s$ to prevent the UAV from getting too close to the unknown space.

\paragraph{Jerk-Matching Penalty}

During replanning, since only $\boldsymbol{o}_0^{[h-1]}$ is specified as the initial state, the $h$-th derivative at the handover between the old and new trajectories is not constrained to match that of the previous trajectory. 
With $h=3$ in this work, the jerk is continuous within each newly generated trajectory, but a jerk jump may still occur at the replanning handover point. 
Such a jerk discontinuity can induce abrupt motion changes of the UAV and lead to harmful oscillations of the FOV. 

To alleviate this effect, we introduce a point-wise jerk-matching penalty at the start point of each newly replanned trajectory, so that its initial jerk remains close to that of the previous trajectory.
We assume that, during each planning step, the UAV is either in a hover state with $\dddot{\boldsymbol{o}}_{\mathrm{old}}=\boldsymbol{0}$ or starts from a point on the prior trajectory.
Therefore, we impose the following point-wise jerk-matching penalty:
\begin{equation}
\mathcal{S}_{\mathrm{jerk}} = w_{\mathrm{jerk}} \left\lVert \dddot{\boldsymbol{o}}_0 - \dddot{\boldsymbol{o}}_{\mathrm{old}} \right\rVert_2^2.
\label{ali:constraint_jerk}
\end{equation}
where $\dddot{\boldsymbol{o}}_0$ is the jerk of the new trajectory at the start point, $\dddot{\boldsymbol{o}}_{\mathrm{old}}$ is the jerk at the same point predicted by the previous trajectory, or $\boldsymbol{0}$ in the hover case, and $w_{\mathrm{jerk}}$ is the corresponding weight.

\subsubsection{Segment-wise Penalty Terms}

For constraints that should be enforced over trajectory segments, we use the constraint function $\mathcal{C}_d(\boldsymbol{o}, \ldots, \boldsymbol{o}^{(h)}) \leq 0$ and convert them into segment-wise penalty terms through numerical integration.
Specifically, each trajectory segment is sampled at $n+1$ points:
\begin{equation}
  \begin{aligned}[c]
    & t_{i,j}=\frac{j}{n}T_i, \\
    & \mathcal{C}_d^{i,j}(\boldsymbol{c},\boldsymbol{T})=\mathcal{C}_d\!\left(\boldsymbol{c}_i^\top\boldsymbol{\beta}(t_{i,j}),\ldots,\boldsymbol{c}_i^\top\boldsymbol{\beta}^{(h)}(t_{i,j})\right), \\
    & \mathcal{I}_d(\boldsymbol{c},\boldsymbol{T})=w_d\sum_{i\in\mathbb{I}}\sum_{j=0}^{n}\frac{T_i}{n}\nu_j\mathcal{L}_1\!\left(\mathcal{C}_d^{i,j}(\boldsymbol{c},\boldsymbol{T})\right), 
    \label{ali:constraint_integration}
  \end{aligned}
\end{equation}
where $w_d$ is the weight of the corresponding term, $\mathbb{I}$ denotes the index set of the trajectory segments over which the penalty is imposed, and $(\nu_0,\nu_1,\ldots,\nu_{n-1},\nu_n)=(1/2,1,\ldots,1,1/2)$.
Adding $\mathcal{I}_d$ to the objective function transforms the corresponding constraint into a penalty over the specified trajectory segment.
The applicable segment set $\mathbb{I}$ for each constraint is specified below.

\paragraph{Safety Constraints}

To ensure the UAV's safety, we constrain the ESDF values evaluated along the trajectory.
For the safe segment $\boldsymbol{S}_s$, the trajectory is required to remain in the \textit{known-safe} space:
\begin{equation}
\mathcal{C}_{\mathrm{ss}}=d_s-E_s(\boldsymbol{p})\leq 0,\qquad \boldsymbol{p}\in\boldsymbol{S}_s.
\label{ali:constraint_ss}
\end{equation}
For the transition segment $\boldsymbol{S}_t$, the trajectory is allowed to enter unknown space while remaining collision-free with respect to known obstacles:
\begin{equation}
\mathcal{C}_{\mathrm{su}}=d_s-E_u(\boldsymbol{p})\leq 0,\qquad \boldsymbol{p}\in\boldsymbol{S}_t.
\label{ali:constraint_su}
\end{equation}

\paragraph{Dynamic Constraints}\label{subsubsection:dynamic_constraints}
Following~\cite{wang2022robust}, we model the UAV dynamics using the flat output $\boldsymbol{o} = [\boldsymbol{p}^\top, \psi]^\top$. 
To ensure dynamic feasibility, we constrain the maximum velocity and body rate along the trajectory.
The velocity and body-rate constraints are written as: 
\begin{equation}
\mathcal{C}_{v}=\|\dot{\boldsymbol{p}}\|_2^2-v_{\max}^2\leq 0,\qquad \boldsymbol{p}\in\boldsymbol{S}_s\cup\boldsymbol{S}_t,
\end{equation}
\begin{equation}
\mathcal{C}_{\omega}=\|\boldsymbol{\omega}\|_2^2-\omega_{\max}^2\leq 0,\qquad \boldsymbol{\omega}\in\boldsymbol{S}_s\cup\boldsymbol{S}_t.
\end{equation}
where $v_{\max}$ is the maximum velocity of the UAV, $\boldsymbol\omega=2(\boldsymbol{q}_z\otimes\boldsymbol{q}_\psi)^{-1}\otimes(\dot{\boldsymbol{q}}_z\otimes\boldsymbol{q}_\psi+\boldsymbol{q}_z\otimes\dot{\boldsymbol{q}}_\psi)$ is the body rate, and $\omega_{\max}$ is the maximum angular rate.

To maintain a stable flight posture during maneuvers, we impose the tilt-angle constraint:
\begin{equation}
\mathcal{C}_{\vartheta}=\arccos\!\left(\boldsymbol{e}_3^\top\boldsymbol{R}(\boldsymbol{q})\boldsymbol{e}_3\right)-\vartheta_{\max}\leq 0,\qquad \boldsymbol{q}\in\boldsymbol{S}_s\cup\boldsymbol{S}_t, 
\end{equation}
where $\vartheta_{\max}$ is the maximum tilt angle.

Finally, the thrust is constrained by: 
\begin{equation}
\mathcal{C}_{f}=(f-f_m)^2-f_r^2\leq 0,\qquad f\in\boldsymbol{S}_s\cup\boldsymbol{S}_t,
\end{equation}
where $f_m=(f_{\max}+f_{\min})/2$ and $f_r=(f_{\max}-f_{\min})/2$, $f_{\max}$ and $f_{\min}$ are the maximum and minimum thrust forces, respectively.

\paragraph{Duration-Balance Penalties}

During the optimization process, the time durations of the segments could become unbalanced.
We aim to keep the time allocated to each segment within a certain deviation range, ensuring that they do not vary excessively from one another.
For the $i$-th segment in $\boldsymbol{S}_s$, we impose the balanced duration constraint:
\begin{equation}
    \begin{aligned}[c]
        & \mathcal C_{T_\mathrm{low}}(T_i)=\epsilon_\mathrm{low}\frac{1}{m_1}\sum_{m=1}^{m_1} T_m - T_i \leq 0, \\ 
        & \mathcal C_{T_\mathrm{upp}}(T_i)=T_i - \epsilon_\mathrm{upp}\frac{1}{m_1}\sum_{m=1}^{m_1} T_m \leq 0, 
        \label{ali:unitime}
    \end{aligned}
\end{equation}
where $i\in[1, \ldots, m_1]$, $\epsilon_{\mathrm{low}} \in (0,1)$ and $\epsilon_{\mathrm{upp}} > 1$ are lower and upper bounds of the allowed deviation, respectively.
The trajectory segments in $\boldsymbol{S}_t$ are constrained in a similar manner.

\subsection{Optimization Problem}
The full optimization problem is formulated as:
\begin{align}
\min_{\overline{\boldsymbol{o}}', \boldsymbol{\tau},  \varrho, \boldsymbol{\zeta}, \psi_u, \dot{\boldsymbol{p}}_f, \psi_f} \mathcal{J} =\ 
&\underbrace{\int_0^{T_A} \boldsymbol{o}^{(h)}(t)^\top \boldsymbol{W} \boldsymbol{o}^{(h)}(t)  dt}_{\text{control effort}} \nonumber \\
&+ \underbrace{\epsilon_{T_s} T_s + \epsilon_{T_u} T_u}_{\text{time regularization}} \nonumber \\
&+ \underbrace{w_\rho \rho(\varrho) + \mathcal{I}_{\mathrm{ap}} + \mathcal{S}_{\mathrm{ap}}}_{\text{active perception}} \nonumber\\
&+ \sum_{d\in \mathbb{D}}\mathcal{I}_d + \mathcal{S}_{\mathrm{jerk}} + \mathcal{S}_{\mathcal{B}s},
\label{ali:final_optimization}
\end{align}
where $\overline{\boldsymbol{o}}' = [\boldsymbol{o}_1', \ldots, \boldsymbol{o}_{m_1-1}', \boldsymbol{o}_{m_1+1}', \ldots, \boldsymbol{o}_{M-1}']$ is obtained from $\boldsymbol{o}'$ in Eq.~\eqref{ali:mincoK} by removing the boundary point $\boldsymbol{o}_{m_1}' \equiv [\boldsymbol{p}_u^\top,\psi_u]^\top$, which is instead parameterized by the unconstrained scalars $\boldsymbol{\zeta}$ in Eq.~\eqref{ali:boundary_reparameterization} and the yaw angle $\psi_u$.
$\boldsymbol{W} \in \mathbb{R}^{4\times4}$ is the weight matrix used to penalize the control effort.
Let \(T_A=\sum_{i=1}^M T_i\) denote the total duration, \(T_s=\sum_{i=1}^{m_1} T_i\) denote the duration of \(\boldsymbol{S}_s\), and \(T_u=\sum_{i=m_1+1}^{M} T_i\) denote the duration of \(\boldsymbol{S}_t\).
We set weights $\epsilon_{T_u} > \epsilon_{T_s}$ to reduce the trajectory duration in the unknown space, thereby driving $\boldsymbol{p}_u$ toward the temporary terminal position $\boldsymbol{p}_{tf}$.
This promotes greater utilization of the known space for maneuvers while minimizing the frequency of active perception, thereby enhancing operational efficiency.
$\mathcal{S}_{\mathcal{B}s} \triangleq  \mathcal{S}_{\mathcal{B}s}^{\min}(\boldsymbol{p}_u) + \mathcal{S}_{\mathcal{B}s}^{\max}(\boldsymbol{p}_u)$ denotes the soft boundary-consistency penalties in Eq.~\eqref{ali:boundary_esdf}.
The subscript $d$ in $\mathcal{I}_d$ indexes the constraints introduced in Sec.~\ref{ssec:constraints}, where $\mathbb{D} = \{\mathrm{ss},\mathrm{su},v,\omega,\vartheta,f,T_\mathrm{low},T_\mathrm{upp}\}$.

\subsection{Replanning}
Trajectory replanning is performed in real-time as the UAV follows the current trajectory, allowing it to promptly respond to newly revealed map information.
At the start of execution, we record the ESDF value \( E_u(\boldsymbol{p}_u) \) of the boundary point \( \boldsymbol{p}_u \) from the previous trajectory. 
During execution, we continuously monitor \( E_s(\boldsymbol{p}_u) \).
If \( E_s(\boldsymbol{p}_u) \) changes, we treat this as a heuristic indication that the boundary neighborhood has been updated by observation, and trigger replanning.
If \( E_s(\boldsymbol{p}_u) \) remains unchanged but \( E_u(\boldsymbol{p}_u) \leq d_s \), the boundary point is near an obstacle, which may cause occlusion and prevent the observation of \( \boldsymbol{p}_v \), thereby triggering replanning.
If by time \( t_{as}+T_p \) at the endpoint \( \boldsymbol{o}_{pf} \) of the AP segment, \( E_s(\boldsymbol{p}_u) \) remains unchanged, we declare the observation a failure and replan immediately.

To account for planner latency, we set a deadline \( T_R \), which equals the expected computation time.
When replanning is triggered at \( t_c \), we designate the handover time at \( t_c + T_R \) and its corresponding state \( \boldsymbol{o}_R \) as the start state of the next replanning.
Provided the planner finishes within \( T_R \) and the new trajectory is delivered by \( t_c + T_R \), the controller seamlessly switches from the old trajectory to the new one at \( \boldsymbol{o}_R \), preserving continuity and smoothness.

If the whole front-end path lies within \( \mathcal{D}_s \), the planner will calculate a trajectory directly to the final goal. 
The trajectory optimization problem is then simplified to a standard MINCO formulation, involving only safety and dynamic constraints.

\section{Simulations}\label{sec:simulations}
In this section, we evaluate \ACRONYM{} with several representative baselines in various environments.

\subsection{LiDAR Sensors}
LiDAR is widely used in robotics due to its stable and reliable perception performance.
Omnidirectional LiDARs, such as those with a full 360$^\circ$ horizontal FOV, provide complete coverage in yaw.
This characteristic eliminates the need for yaw planning in observation trajectories, allowing the optimizer to focus primarily on vertical motion.
Extensive benchmarking in diverse environments demonstrates the effectiveness of \ACRONYM{}.

\subsubsection{Horizontal Narrow-Space Scenario}\label{sssec:sim_radar_horrizon}

We design a horizontally winding corridor to evaluate \ACRONYM{} during horizontal flight using an omnidirectional LiDAR.
The winding corridor introduces frequent obstacle occlusions, and aggressive maneuvers may induce large attitude tilts that increase collision risk under a limited vertical FOV.
Therefore, we evaluate \ACRONYM{} under different vertical FOV settings to examine its effectiveness in achieving safe and efficient navigation in horizontally constrained environments.

We compare \ACRONYM{} with two baseline approaches.
SUPER~\cite{ren2025safety} adopts a dual-trajectory strategy consisting of a high-speed exploration trajectory and a safety-guaranteed backup trajectory.
We further consider a MINCO-based implementation of FASTER~\cite{tordesillas2021faster,wang2022geometrically}, denoted as FM, to provide a fair comparison by unifying the trajectory representation across methods.
Specifically, FM preserves FASTER's dual-trajectory strategy. 
It first optimizes an exploration trajectory $\Gamma_e$ by allowing traversal through unknown space, and then identifies the point $\boldsymbol{p}_u$ where $\Gamma_e$ first enters the unknown region.
Starting from $\boldsymbol{p}_u$, a forward search is performed along $\Gamma_e$ to find the first point $ \boldsymbol{p}_{\mathrm{safe}} $ satisfying the safety condition in Eq.~\eqref{ali:Csd}.
A backup trajectory $\Gamma_s$ constrained in known free space is then planned from $ \boldsymbol{p}_{\mathrm{safe}} $, with its terminal state jointly optimized following Sec.~\ref{ssec:truncation_and_relaxation}.
To account for the limited vertical FOV in this LiDAR setting, FM further incorporates an explicit tilt-angle constraint based on the differential-flatness formulation of MINCO, so that the future horizontal flight trajectory remains within the observable vertical range and obstacles do not become unobservable due to excessive attitude tilt \cite{lu2021flight, lu2022real}.
In this way, FM serves as a risk-aware baseline that addresses the vertical-FOV-induced safety issue without introducing explicit active perception.

\begin{table}[h]
    \centering
    \caption{
        Comparison of \ACRONYM{}, SUPER, and FM.
    }
    \setlength{\extrarowheight}{2pt} 
    \setlength{\tabcolsep}{4.8pt}
\resizebox{\linewidth}{!}{
\begin{tabular}{c|ccccccc}
\toprule
  \begin{tabular}[c]{@{}c@{}}Vertical\\FOV\end{tabular} &          & \begin{tabular}[c]{@{}c@{}}TD\\ $s$\end{tabular} & \begin{tabular}[c]{@{}c@{}}TL\\ $m$\end{tabular} & \begin{tabular}[c]{@{}c@{}}E\\ $m^2/s^5$\end{tabular} & \begin{tabular}[c]{@{}c@{}}MV\\ $m/s$\end{tabular} & Result      & \begin{tabular}[c]{@{}c@{}}P\\ AP/BT\end{tabular}        \\ \midrule
\multirow{3}{*}{90$^\circ$}  & \ACRONYM{}    & 40.78      & 72.22      & 138.01     & 1.77  & \textbf{T} 10/10   & 27.73\%  \\
                     & SUPER    & 41.85      & 81.44      & 171.42     & 1.95               & \textbf{T} 10/10   & 0.02\%   \\
                     & FM       & 44.42      & 69.42      & 137.50     & 1.56               & \textbf{T} 10/10   & 1.72\%   \\ \midrule
\multirow{3}{*}{30$^\circ$}  & \ACRONYM{}    & 51.81      & 77.38      & 188.01     & 1.49  & \textbf{T} 10/10   & 31.86\%  \\
                     & SUPER    & -          & -          & -          & -                  & \textbf{F} 79.19\% & -        \\
                     & FM       & 55.09      & 73.56      & 158.34     & 1.34               & \textbf{T} 10/10   & 4.94\%   \\ \midrule
\multirow{3}{*}{10$^\circ$}  & \ACRONYM{}    & 69.85      & 88.06      & 170.76     & 1.26  & \textbf{T} 10/10   & 32.56\%  \\ 
                     & SUPER    & -          & -          & -          & -                  & \textbf{F} 66.04\% & -        \\
                     & FM       & 85.68      & 76.27      & 162.23     & 0.89               & \textbf{T} 10/10   & 3.39\%   \\ \midrule
\multirow{3}{*}{0.2$^\circ$} & \ACRONYM{}    & 155.55     & 115.22     & 288.70     & 0.74  & \textbf{T} 6/10    & 40.72\%  \\
                     & SUPER    & -          & -          & -          & -                  & \textbf{F} 39.74\% & -        \\
                     & FM       & -          & -          & -          & -                  & \textbf{F} Infeasible        & -        \\ \bottomrule
\end{tabular}
}
\vspace{-0.2cm}
\label{table:super_horizon}
\end{table}

We summarize the performance metrics in Table~\ref{table:super_horizon}: Trajectory Duration (TD, in seconds), Trajectory Length (TL, in meters), Energy (E, defined as the integral of squared jerk in $m^2/s^5$), Mean Velocity (MV, in $m/s$), Result (reported as \textbf{T} $N/10$ when at least one trial succeeds, where $N/10$ denotes the success rate over 10 trials; otherwise reported as \textbf{F} $x\%$, where $x\%$ denotes the maximum completion ratio), and the Proportion (P) of active perception (AP) or backup trajectory (BT) usage relative to the total trajectory length.
For settings with partial success, TD, TL, E, MV, and P are averaged over successful trials only.

\begin{figure}[t]
  \centering
  \includegraphics[width=\linewidth]{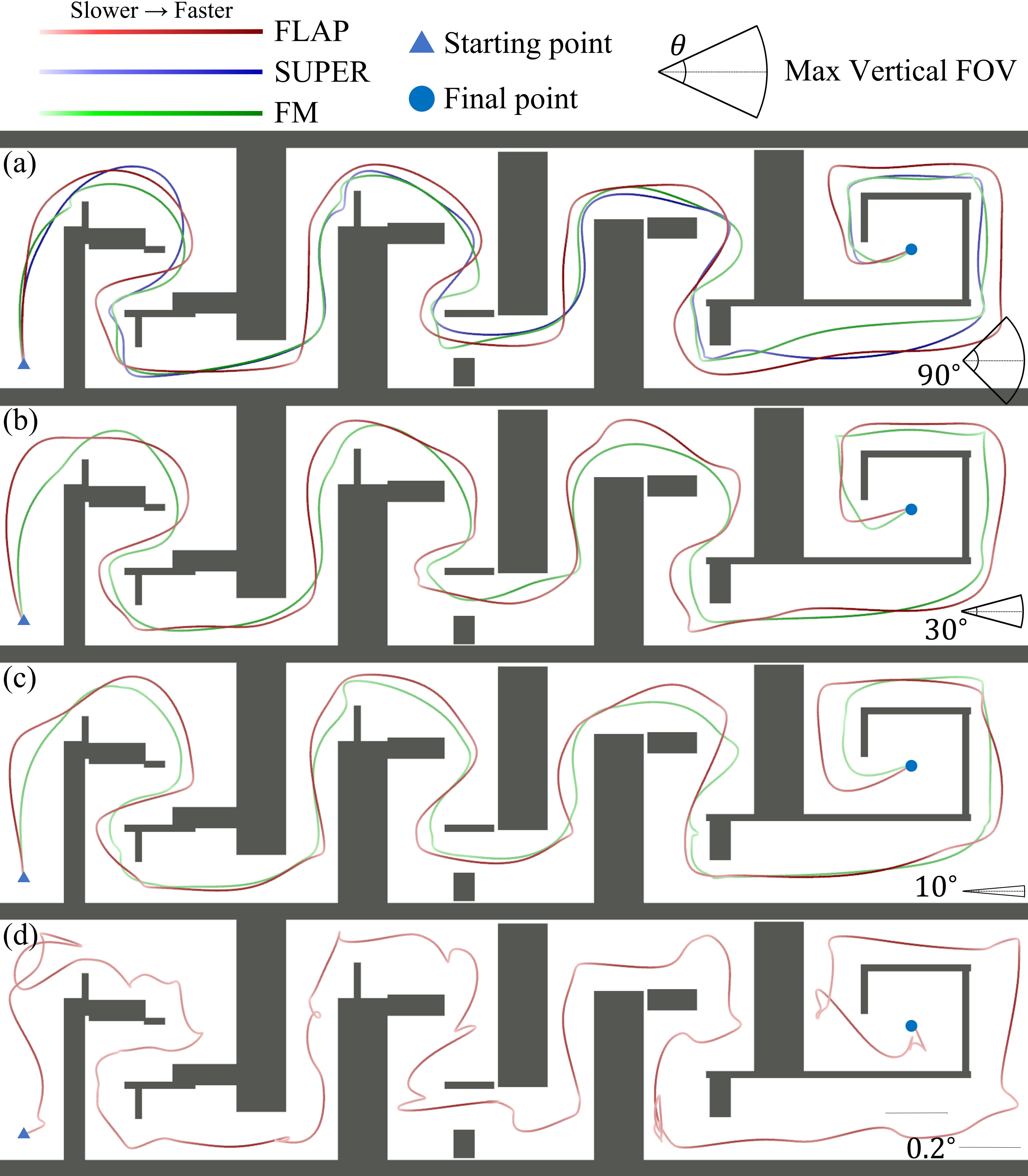}
  \caption{
    Results of \ACRONYM{}, SUPER, and FM in the horizontal narrow-space scenario.
    We use gradient colors to represent the UAV's speed; the darker the color, the higher the speed.
    From top to bottom are the results with total vertical FOVs of 90$^\circ$, 30$^\circ$, 10$^\circ$, and 0.2$^\circ$.
    The small fan-shaped inset at the right of each row illustrates the corresponding vertical sensing range.
    The successful trajectory is shown only if the method succeeds in at least one trial.
    No trajectory is shown when all trials fail.
}
  \label{fig:super_horizon}
  \vspace{-0.5cm}
\end{figure}

When the vertical FOV is 90$^\circ$, even at maximum velocity, the UAV's flight direction remains within the sensor's FOV. 
In this case, all three methods achieve comparable performance, as shown in Fig.~\ref{fig:super_horizon}(a).
When the vertical FOV is reduced to 30$^\circ$, the situation changes. 
During high-speed flight, the UAV's tilt causes the sensor's FOV to fall below the horizontal plane along the flight direction.
This limitation has the most severe impact on SUPER, which completes less than 80\% of the trajectory in all trials. 
A likely explanation is that SUPER's front-end search path lies largely outside the FOV during high-speed segments, which prevents successful optimization of the backup trajectory.
In contrast, FM completes the task by imposing stricter acceleration and velocity limits through maximum tilt angle constraints, though at the cost of increased mission time.
\ACRONYM{} ensures safety through active perception, as illustrated in Fig.~\ref{fig:super_horizon}(b).
Despite generating longer trajectory lengths due to detours for observation, it achieves higher overall efficiency by maintaining greater average velocities.
However, because the front-end adopts an optimistic assumption about unknown spaces, sudden braking may occur when previously occluded obstacles are newly observed, resulting in higher energy consumption.

When the vertical FOV is further reduced to 10$^\circ$, \ACRONYM{} demonstrates a more pronounced advantage in operational efficiency, as shown in Fig.~\ref{fig:super_horizon}(c).
When the vertical FOV is reduced to an extreme 0.2$^\circ$, making the LiDAR effectively planar, FM can no longer generate feasible trajectories because the resulting tilt-angle constraint becomes overly restrictive, while \ACRONYM{} still achieves partial success in 6 out of 10 trials, as illustrated in Fig.~\ref{fig:super_horizon}(d).
The reason is that the extremely narrow vertical FOV implies an excessively strict tilt-angle bound; once the UAV starts to move, this bound is readily violated, and FM can no longer find a feasible solution under the resulting constraints.
\ACRONYM{} actively constrains the UAV attitude during the AP segment to observe target locations. 
However, the extremely limited vertical FOV causes the distribution of unknown regions to become highly fragmented, which introduces errors in boundary estimation. 
These errors may lead to incorrect observations, while the strict FOV constraints can also cause optimization failures, triggering emergency stops and resulting in mission failure in some cases.

\subsubsection{Overhead-Obstacle Scenario}\label{sssec:sim_radar_overhead}
To evaluate the performance of \ACRONYM{} in large-scale vertical motion along the z-axis, we design an H-shaped obstacle.
The UAV is required to detect the overhead obstacle and navigate to the top of this structure. 
The vertical FOV of the LiDAR is configured to $\pm 45^\circ$. 
\begin{figure}[t]
  \centering
  \includegraphics[width=\linewidth]{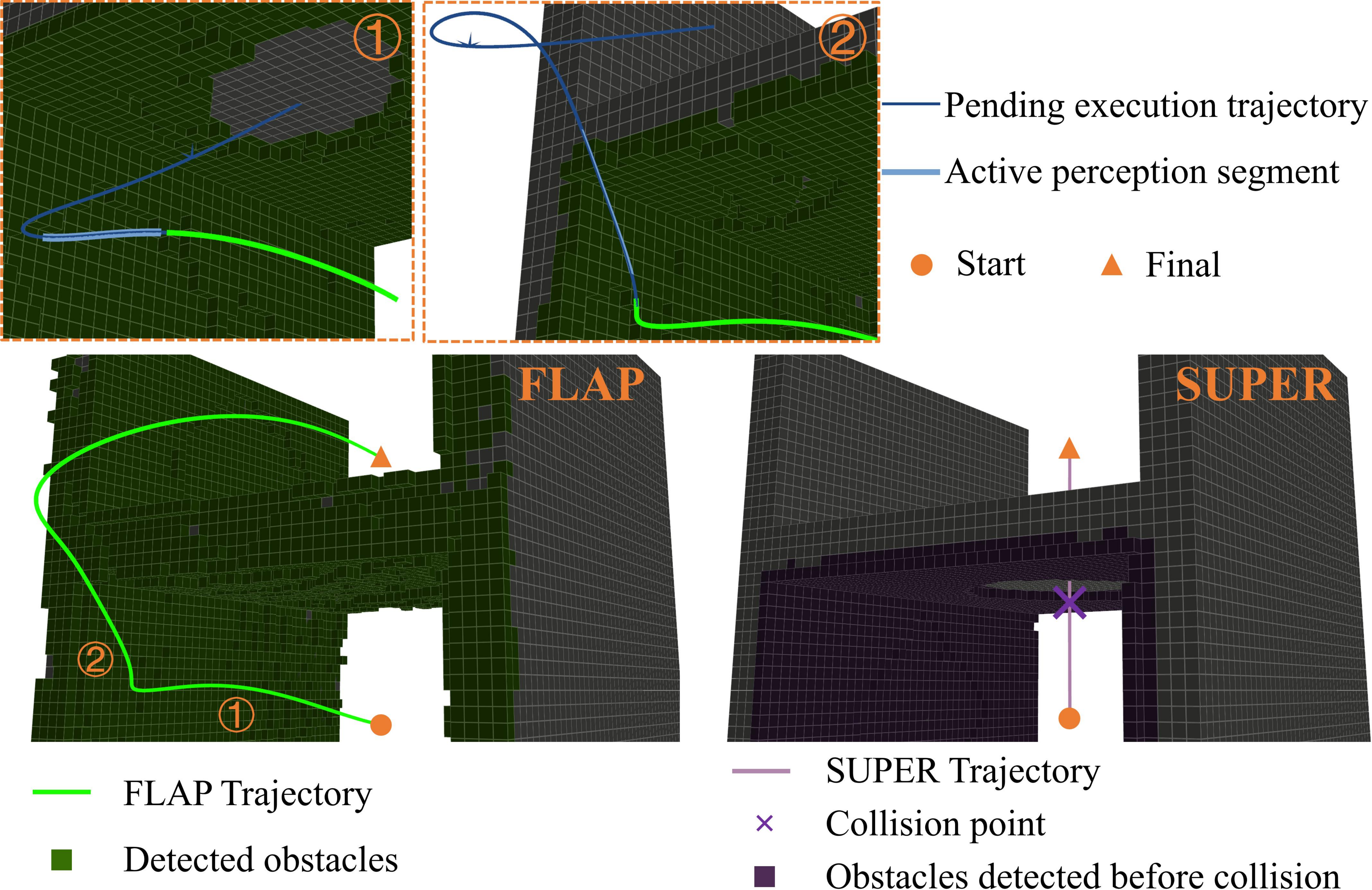}
  \caption{
    Results of \ACRONYM{} and SUPER in the overhead-obstacle scenario. 
    The final point is set directly above the start point, and the UAV must actively observe the obstacle above to ensure safety.
  }
  \label{fig:super_h}
  \vspace{-0.4cm}
\end{figure}

\begin{figure*}[t]
  \centering
  \includegraphics[width=\textwidth]{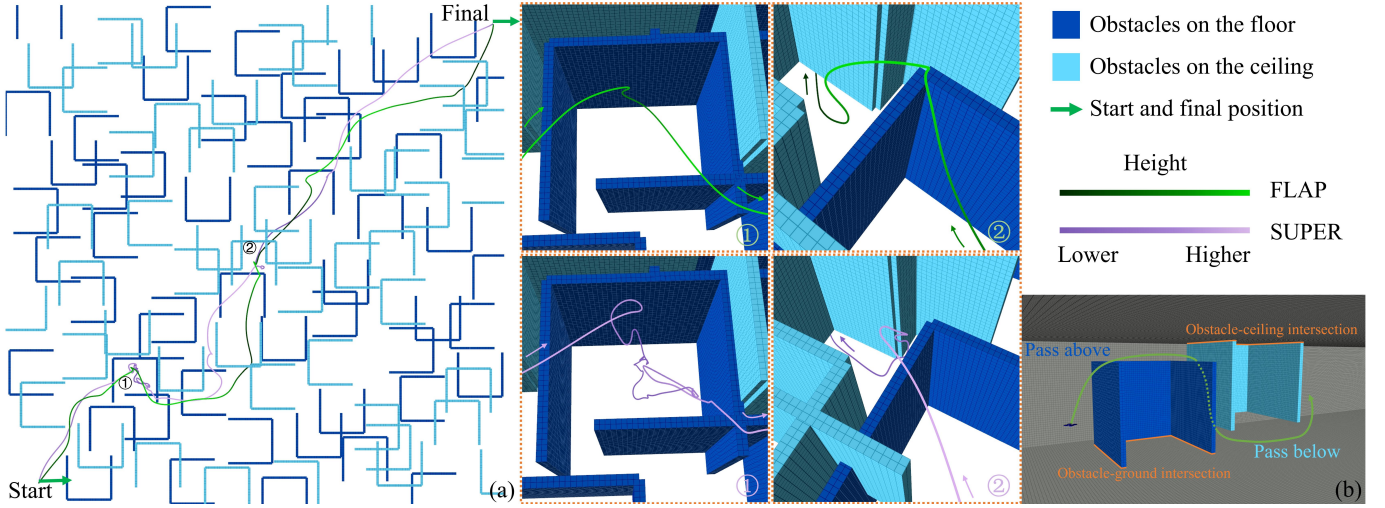}
  \caption{
      Results of \ACRONYM{} and SUPER in the U-shaped maze scenario. 
      (a) Trajectories of the two methods, where color gradients encode the UAV altitude; the right side shows representative close-up views at two locations. 
      (b) Two types of U-shaped obstacles in the environment, where two colors distinguish obstacles attached to the floor and the ceiling, and the orange lines mark their intersections with the ground or ceiling. 
    }
  \label{fig:super_umaze}
  \vspace{-0.4cm}
\end{figure*}

SUPER's backup trajectory operates within the safety space. 
This approach, in practice, necessitates a LiDAR system with omnidirectional coverage to ensure the validity of the free-space assumption.  
However, when moving along the z-axis, the initial trajectory may lie entirely in the unknown space. 
As a result, SUPER fails to generate a valid backup trajectory, leading to misjudgment of obstacles within its FOV blind zone, as shown in Fig.~\ref{fig:super_h}.

In contrast, \ACRONYM{} actively explores unknown spaces through active perception. 
The trajectory first maneuvers within the known space to reach a vantage point suitable for observing the region above, as shown at \ding{172} in Fig.~\ref{fig:super_h}. 
When the most direct vertical path is found to be obstructed, the trajectory deviates and continues to ascend within known spaces at \ding{173}, while simultaneously gathering observations of the unknown space toward the target. 
\ACRONYM{} enables a spiral ascent pattern. 
Throughout this motion, the method promotes thorough observation of potentially reachable goal locations, thereby enhancing trajectory safety.

\subsubsection{U-Shaped Maze Scenario}\label{sssec:sim_radar_umaze}

We further design a complex U-shaped maze scenario to evaluate the performance of both methods in a more complex environment, as shown in Fig.~\ref{fig:super_umaze}.
The environment contains two types of U-shaped obstacles: one suspended from the ceiling that requires the UAV to pass from below, and another placed on the ground that requires passage from above, as shown in Fig.~\ref{fig:super_umaze}(b).
To traverse these obstacles, the UAV must execute coupled maneuvers in the xy-plane and along the z-axis to reach the goal.
In this scenario, the LiDAR is configured with a vertical FOV of $\pm 45^\circ$.

\ACRONYM{} performs effectively in this U-shaped maze scenario. 
For obstacles on the ground, the UAV ascends to avoid them, as shown at \ding{172} in Fig.~\ref{fig:super_umaze}. 
At higher altitudes facing suspended obstacles, the UAV executes a spiral descent to maintain safety while exploring feasible paths, ultimately discovering the exit as shown at \ding{173} in Fig.~\ref{fig:super_umaze}.

In contrast, SUPER's backup trajectory, designed for safety assurance, exhibits limitations when facing tall obstacles. 
Although its exploratory trajectory actively explores different altitudes, the backup trajectory conservatively outputs paths confined to known space. 
At \ding{172} in Fig.~\ref{fig:super_umaze}, since bypassing from one side would move away from the goal, the exploratory trajectory tends to explore whether passable regions exist below. 
However, because the trajectory stays close to the obstacle, SUPER remains locked in backup trajectory execution, failing to complete exploration of the lower region. 
By contrast, \ACRONYM{} quickly completes exploration and chooses to bypass the obstacle. 
At \ding{173}, SUPER exhibits similar behavior, yet because the space is relatively spacious, it can discover a feasible path below after hovering for a while.

\subsection{Camera Sensors}

Unlike LiDAR sensors, which often provide a wide or even omnidirectional sensing range, vision sensors such as monocular and stereo cameras are typically limited by a narrower forward FOV and a shorter effective depth perception range.
These limitations make trajectory planning more challenging, because the planner must ensure dynamic feasibility and safety while actively adjusting the attitude and position to keep critical unknown regions within the camera FOV and avoid blind areas.
Consequently, in vision-based navigation systems, it is crucial to seamlessly integrate active perception into the trajectory optimization process, especially when handling the coupled constraints among pitch, yaw, and position in 3D space.
This section evaluates the performance of \ACRONYM{} under typical vision sensor configurations and validates its adaptability and robustness in the presence of limited FOV and complex environments.

\subsubsection{Horizontal Narrow-Space Scenario}

\begin{figure*}[t]
    \centering
    \subfloat[
      Trajectory comparison of \ACRONYM{}, RAPTOR, FM and NBV in the horizontal narrow-space scenario. 
      The UAV is equipped with a forward-facing depth camera with an approximately 78$^\circ$ horizontal FOV. 
      Markers \ding{172}–\ding{177} denote representative locations for detailed analysis.
    \label{fig:camera_hori}]{
        \includegraphics[width=\textwidth]{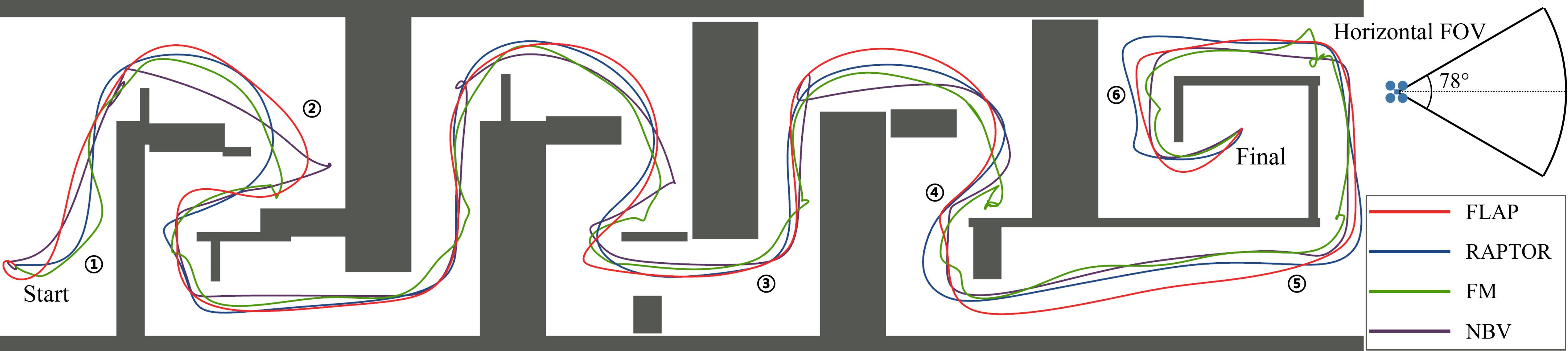}
    }\\[3pt]
    \vspace{-0.3cm}
    \subfloat[
      Representative close-up views at the locations marked in (a), illustrating the horizontal visibility during navigation, together with the optimized trajectory, executed trajectory, AP segment of \ACRONYM{}, and backup trajectory of FM.
      For \ACRONYM{}, FM, and NBV, the purple regions denote detected obstacles.
      For RAPTOR, the purple regions denote inflated obstacles used for planning, while the original map obstacles are shown in gray.
    \label{fig:camera_hori_zoom}]{
        \includegraphics[width=\textwidth]{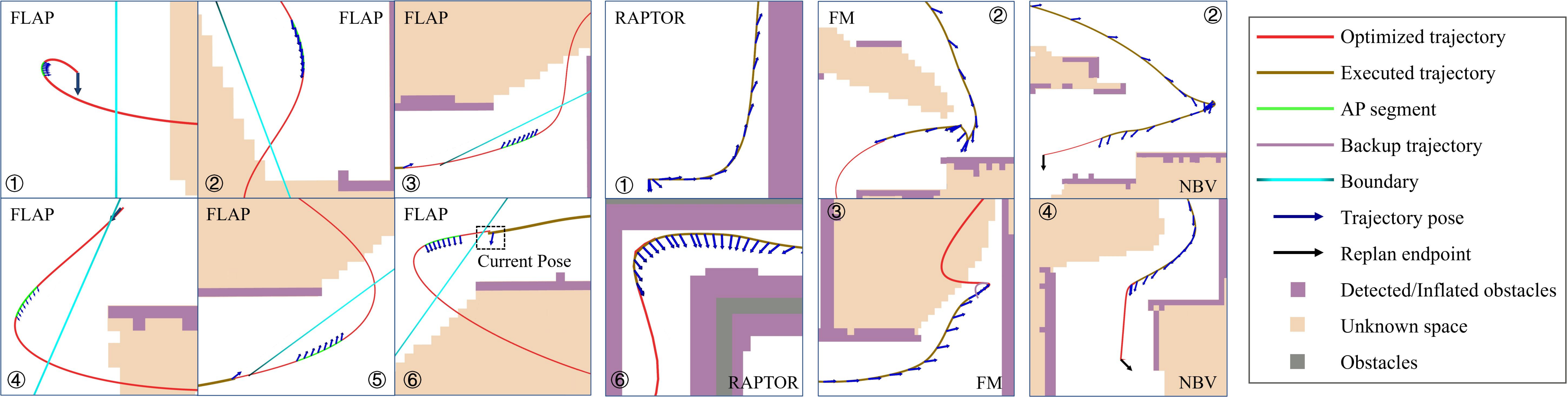}
    }
    \caption{
      Comparison of \ACRONYM{}, RAPTOR, FM, and NBV in the horizontal narrow-space scenario.
    }
    \vspace{-0.4cm}
  \label{fig:camera_hori_comp}
\end{figure*}

We use the same horizontal environment as in Sec.~\ref{sssec:sim_radar_horrizon} to evaluate \ACRONYM{}. 
To restrict the UAV's vertical motion, the floor and ceiling heights of the environment are set to approach each other. 
The depth camera, aligned with the UAV's heading direction, is used as the perceptual input, requiring the UAV to actively plan its heading angle to observe the environment. 
We set the UAV's maximum velocity to $3\,\mathrm{m/s}$.

Typical methods based on environmental assumptions, such as pessimistic strategies that avoid trajectories entering unknown space, often rely on next-best-view or frontier-based viewpoint selection to choose intermediate observation goals in known free space \cite{connolly1985determination, liu2016high}.
When the final goal lies in unknown space, such methods typically select a frontier-associated observation viewpoint and use it as a temporary goal, so that new regions can be observed.
To mitigate the influence of trajectory parameterization on trajectory quality and computational cost, we also implement Next-Best-View (NBV) within the MINCO framework~\cite{wang2022geometrically}.
Specifically, the weighted A* algorithm from Sec.~\ref{ssec:front_end} is used to obtain the global path and identify the first point \( \boldsymbol p_u \) entering the unknown space. 
For a fair comparison with the proposed formulation, we compute the target observation point \( \boldsymbol p_v \) using the same construction as in Sec.~\ref{subsec:visibility_criteria}.
Centered at \( \boldsymbol p_v \), we perform discrete sampling along the horizontal and vertical directions to obtain candidate viewpoints \( \boldsymbol p_f^{\text{NBV}} \) from which \(\boldsymbol p_v \) is observable. 
We select the nearest non-occluded candidate to the start as the final position and set the final yaw to the horizontal component of the direction from \( \boldsymbol p_f^{\text{NBV}} \) to \( \boldsymbol p_v \).

As a representative risk-aware baseline, we use the same FM baseline introduced in Sec.~\ref{sssec:sim_radar_horrizon}.
Different from the LiDAR setting, the camera has a limited horizontal FOV and therefore requires explicit heading planning.
We additionally optimize the yaw angle by imposing a segment-wise penalty term.
The corresponding penalty is imposed on both $\Gamma_e$ and $\Gamma_s$:
\begin{equation}
\mathcal{P}_{\mathrm{ori}} = \|\dot{\boldsymbol{p}}_{xy}\|_2 - \dot{\boldsymbol{p}}_{xy}^{\top}\boldsymbol{\Theta},
\qquad
\boldsymbol{\Theta} = [\cos\psi,\sin\psi]^{\top},
\end{equation}
where $\dot{\boldsymbol{p}}_{xy} = [\dot p_x,\dot p_y]^{\top}$ is the horizontal velocity and $\psi$ denotes the UAV heading angle.
This penalty encourages the heading angle to remain aligned with the velocity direction for forward observation.

RAPTOR~\cite{zhou2021raptor} further incorporates risk-aware trajectory refinement that considers information acquisition.
It constrains the distance from the latest braking point to the line connecting the current position and the obstacle, thus enhancing safety by minimizing occlusion.
Active yaw planning is also integrated to enhance both safety and navigation efficiency in unknown environments.

As shown in Fig.~\ref{fig:camera_hori_comp}(a), we evaluate four methods in the horizontal narrow-space scenario. 
At \ding{172}, if the UAV's heading deviates substantially from its direction of motion, \ACRONYM{} utilizes the initialized known space to perform a brief backward maneuver. 
This allows it to safely rotate its yaw toward the target space for observation within the AP segment.
At \ding{173}-\ding{177}, the UAV consistently rotates its heading toward the unknown spaces. 
The preceding AP segment promotes earlier observation of unknown spaces without sacrificing smoothness or efficiency, resulting in smoother trajectories.

NBV ensures safety under a pessimistic assumption, but its conservative policy can lead the UAV to observe partially ineffective waypoints. 
Because the final velocity must be reduced to zero for safety, the trajectory undergoes frequent acceleration-deceleration, especially at corners such as \ding{173}, which reduces efficiency.

FM also ensures safety through dual trajectories with distinct roles. 
However, since the objective of the backup trajectory is safety rather than active perception, the UAV can become trapped in a confined local region in narrow environments.
For instance, at \ding{173}, the UAV pauses in front of an obstacle before finding the way out; at \ding{175}, the exploratory trajectory tends left while the backup trajectory requires a rightward detour, and proximity to unknown spaces causes the UAV to execute the rightward backup trajectory almost continuously. 
With the heading constrained to align with the velocity direction, the brief execution of the exploratory trajectory still leaves insufficient time to rotate the yaw for observation.

RAPTOR incorporates obstacle avoidance and perception-aware refinement, but it does not explicitly impose the visibility constraints.
In particular, since the heading is planned independently, the UAV may fail to reorient toward the region to be observed in time, as shown at \ding{172} in Fig.~\ref{fig:camera_hori_comp}(a). 
When the initial position is close to an obstacle and the initial yaw is opposite to the desired observation direction, insufficient time may be reserved for yaw reorientation during forward flight, causing RAPTOR to collide before the obstacle enters the camera FOV. 
In the subsequent segments, RAPTOR also actively steers the heading toward unknown regions, showing a perception-oriented behavior similar to \ACRONYM{}.

\begin{table}[h]
    \centering
    \vspace{-0.2cm}
    \caption{
        Comparison of \ACRONYM{}, NBV, FM and RAPTOR.
    }
    \setlength{\extrarowheight}{2pt} 
\begin{tabular}{c|cccccc}
\toprule
         & \begin{tabular}[c]{@{}c@{}}TD\\ $s$\end{tabular} & \begin{tabular}[c]{@{}c@{}}TL\\ $m$\end{tabular} & \begin{tabular}[c]{@{}c@{}}E\\ $m^2/s^5$\end{tabular} & \begin{tabular}[c]{@{}c@{}}SSAY\\ $rad^2/s^4$\end{tabular} & \begin{tabular}[c]{@{}c@{}}CT\\ $ms$\end{tabular} & RP  \\ \midrule
\ACRONYM{}    & 40.17                                            & 73.37                                            & 165.98                                                     & 121.10                                                    & 5.49                                              & 182 \\
NBV      & 74.51                                            & 73.12                                            & 112.04                                                     & 84.78                                                     & 0.88                                              & 231 \\
FM       & 88.10                                            & 81.97                                            & 188.49                                                     & 45.10                                                     & 2.73                                              & 563 \\
RAPTOR   & 48.43                                            & 72.40                                            & 148.22                                                     & 701.58                                                    & 8.50                                              & 163 \\ \bottomrule
\end{tabular}
\vspace{-0.2cm}
\label{table:camera_horizon}
\end{table}

We also summarize performance metrics in this environment, as shown in Table~\ref{table:camera_horizon}. 
Benefiting from its active perception strategy, \ACRONYM{} reaches the goal with reduced trajectory duration while maintaining relatively high speed and moderate energy consumption.
Although the required observation detours increase the trajectory length, they allow the UAV to preserve motion efficiency by avoiding excessive slowdowns.
Because \ACRONYM{} explicitly plans yaw and penalizes its lack of smoothness, the sum of squared yaw accelerations (SSAY) is lower than RAPTOR; in contrast, FM and NBV primarily benefit from slower motion, which allows more time to rotate yaw. 
The average computation time (CT) remains suitable with real-time execution.
The non-fixed-rate replanning scheme further prevents excessive replanning steps (RP).

\subsubsection{Overhead-Obstacle Scenario}

To evaluate \ACRONYM{}'s effectiveness during large-scale vertical motion along the z-axis with a limited vertical FOV, we conduct experiments in the environment introduced in Sec.~\ref{sssec:sim_radar_overhead}.

\begin{figure}[b]
  \centering
  \vspace{-0.2cm}
  \includegraphics[width=\linewidth]{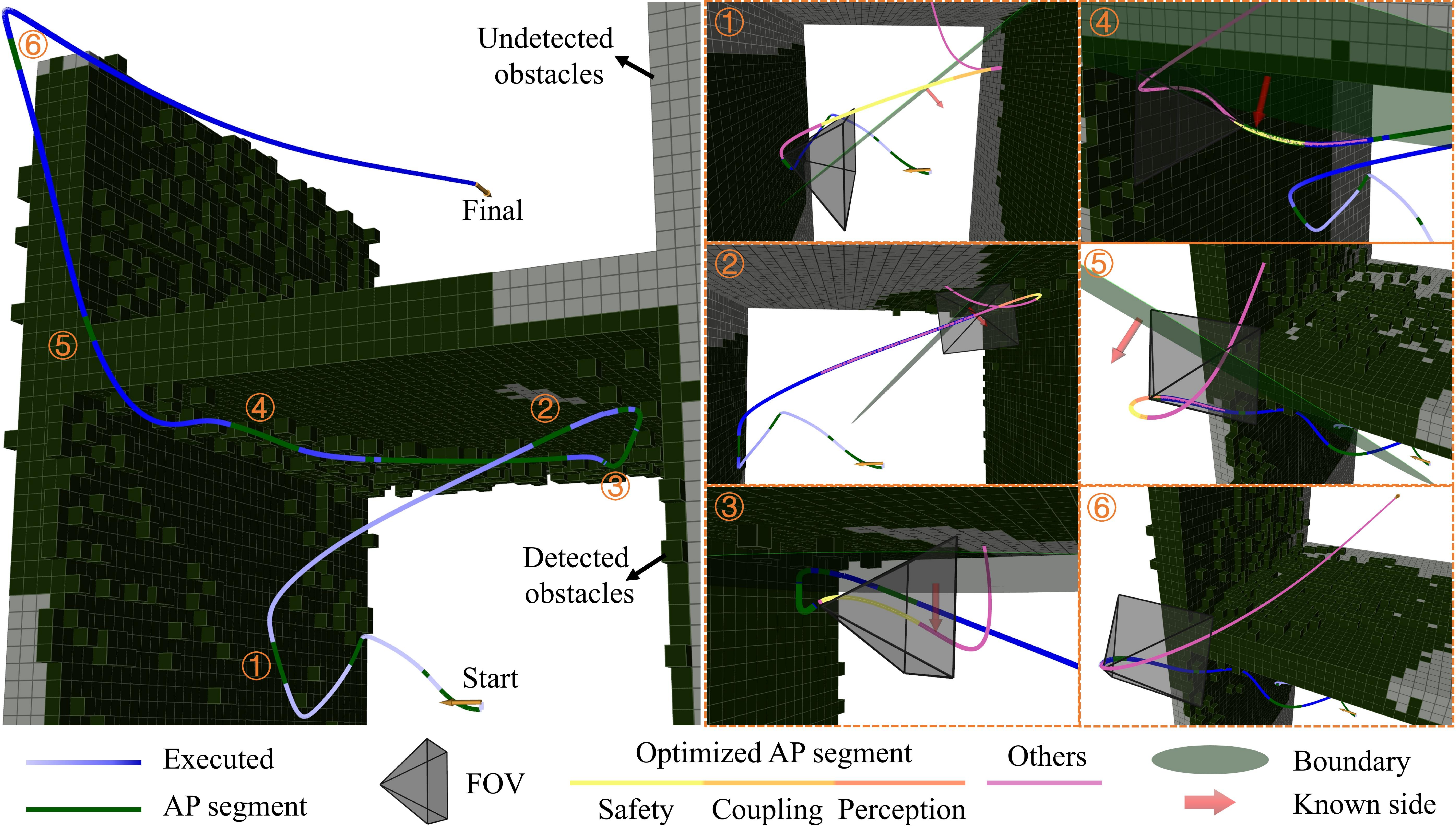}
  \caption{
        Results of \ACRONYM{} in the overhead-obstacle scenario. 
        The left part shows the executed trajectory along with its AP segment. 
        The right part provides zoomed-in views at several locations, illustrating the optimized trajectory and the UAV's current FOV.
        The AP segment can be interpreted as containing Safety, Coupling, and Perception phases induced by the activation mechanism.
    }
  \label{fig:camera_proposed_h}
\end{figure}

As shown in Fig.~\ref{fig:camera_proposed_h}, the UAV must observe the overhead obstacle and navigate to the top of the structure.
The optimized trajectory consists of the AP segment and the remaining part of the trajectory.
According to the activation function in Eqs.~\eqref{ali:La} and \eqref{ali:Cap}, the AP segment can be interpreted as consisting of three phases, namely Safety, Coupling, and Perception.
These phases are not manually predefined, but emerge from the optimized trajectory under the proposed risk-aware formulation.
In the Safety, Coupling, and Perception parts, the activation function satisfies $\mathcal{L}_a(\mathcal{C}_{\mathrm{sd}})=0$, $0<\mathcal{L}_a(\mathcal{C}_{\mathrm{sd}})<1$, and $\mathcal{L}_a(\mathcal{C}_{\mathrm{sd}})=1$, respectively, corresponding to inactive, partially activated, and fully imposed visibility penalties.

\begin{figure}[t]
  \centering
  \includegraphics[width=\linewidth]{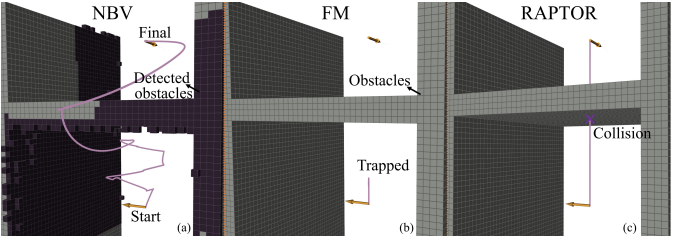}
  \caption{
    Results of NBV (a), FM (b), and RAPTOR (c) in the overhead-obstacle scenario. 
    NBV reaches the goal conservatively, FM gets stuck below the obstacle, and RAPTOR collides during aggressive ascent.
    }
  \label{fig:camera_others_h}
  \vspace{-0.3cm}
\end{figure}

At \ding{172} of Fig.~\ref{fig:camera_proposed_h}, the boundary is formed by previously acquired observations.
After the UAV rotates its heading, most of the subsequent trajectory lies on the unknown side of the boundary. 
Consequently, the trajectory first satisfies the \textit{safety criterion} by moving toward the safe side before perception. 
This early placement of the AP segment, enabled by optimizing $\varrho$, reduces the need for abrupt slowing later in the trajectory.
Since the goal lies directly above the obstacle and the weighted A* front-end adopts an optimistic estimate of the unknown space, it initially generates a reference path that passes over the unknown obstacle. 
\ACRONYM{} thus performs sufficient observation of the obstacle, as shown at \ding{173} and \ding{174} in Fig.~\ref{fig:camera_proposed_h}. 
Once observation reveals that overflying the obstacle is infeasible, the UAV selects a lateral bypass, as shown at \ding{175} and \ding{176} in Fig.~\ref{fig:camera_proposed_h}. 
After completing the necessary perception, the UAV gains a clear sight of the goal at \ding{177} in Fig.~\ref{fig:camera_proposed_h} and plans the final trajectory.

We also evaluate NBV, FM, and RAPTOR in this environment, as shown in Fig.~\ref{fig:camera_others_h}.
Benefiting from the safety guarantees of the pessimistic assumption, NBV reaches the goal safely. 
However, due to the limited FOV, upward perception is highly inefficient.
Frequent accelerations and decelerations yield a polyline-like trajectory, resulting in inefficiency and poor smoothness.
The known space initialized around the starting position limits the maximum attainable height of FM. 
This is because FM's aggressive exploratory trajectory persistently attempts a direct vertical ascent into the unknown space, causing the system to execute an almost stationary backup trajectory repeatedly.
As a result, it becomes stuck. 
RAPTOR's planned yaw and constrained distance do not ensure vertical observability, causing the UAV to pursue an overly aggressive trajectory and collide.

\subsubsection{U-Shaped Maze Scenario}

\begin{figure}[t]
  \centering
  \includegraphics[width=\linewidth]{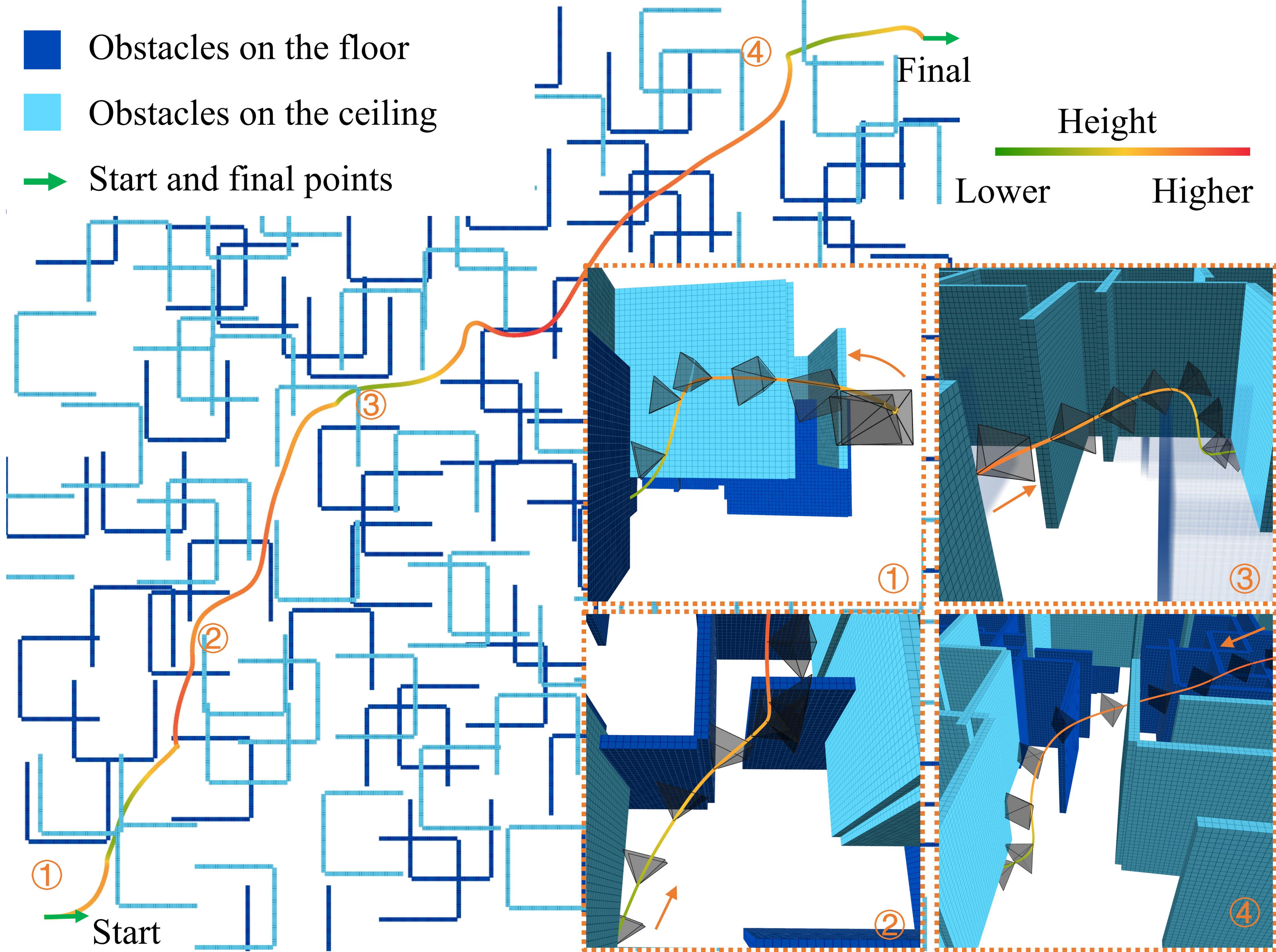}
  \caption{
    Performance of \ACRONYM{} in the U-shaped maze scenario using a depth camera. 
    The trajectory is color-coded by height, with the UAV's orientation marked along the trajectory.
  }
  \label{fig:camera_umaze}
  \vspace{-0.4cm}
\end{figure}

To evaluate the performance of \ACRONYM{} in dense obstacle environments, we also design a scenario featuring multiple U-shaped obstacles, as described in Sec.~\ref{sssec:sim_radar_umaze}. 
Different from the LiDAR setting, when using a camera sensor, the UAV must not only avoid obstacles in 3D space but also actively adjust its yaw angle to keep the unknown space within the camera's FOV.

As shown in Fig.~\ref{fig:camera_umaze}, \ACRONYM{} operates effectively in environments that require complex three-dimensional maneuvers, even when using a sensor with a limited horizontal FOV. 
At \ding{172}, \ding{174}, and \ding{175}, the UAV descends to navigate around tall obstacles detected on both sides. 
At \ding{173}, it ascends to bypass a lower obstacle.
Notably, because the UAV has previously observed and stored obstacle information at current heights during its earlier trajectory, it tends to execute significant vertical maneuvers.
For instance, at \ding{173}, because the right side appears shorter and more direct toward the goal, the UAV decelerates and rotates its yaw to inspect that direction for potential obstructions.
It resumes forward flight only after confirming the presence of the tall obstacle. 
Similarly, at \ding{174} and \ding{175}, because the UAV's view of the terrain below is blocked, it first descends until a safe region is observed before continuing its flight.

\subsubsection{Vertical Pipeline}

\begin{figure}[t]
  \centering
  \includegraphics[width=\linewidth]{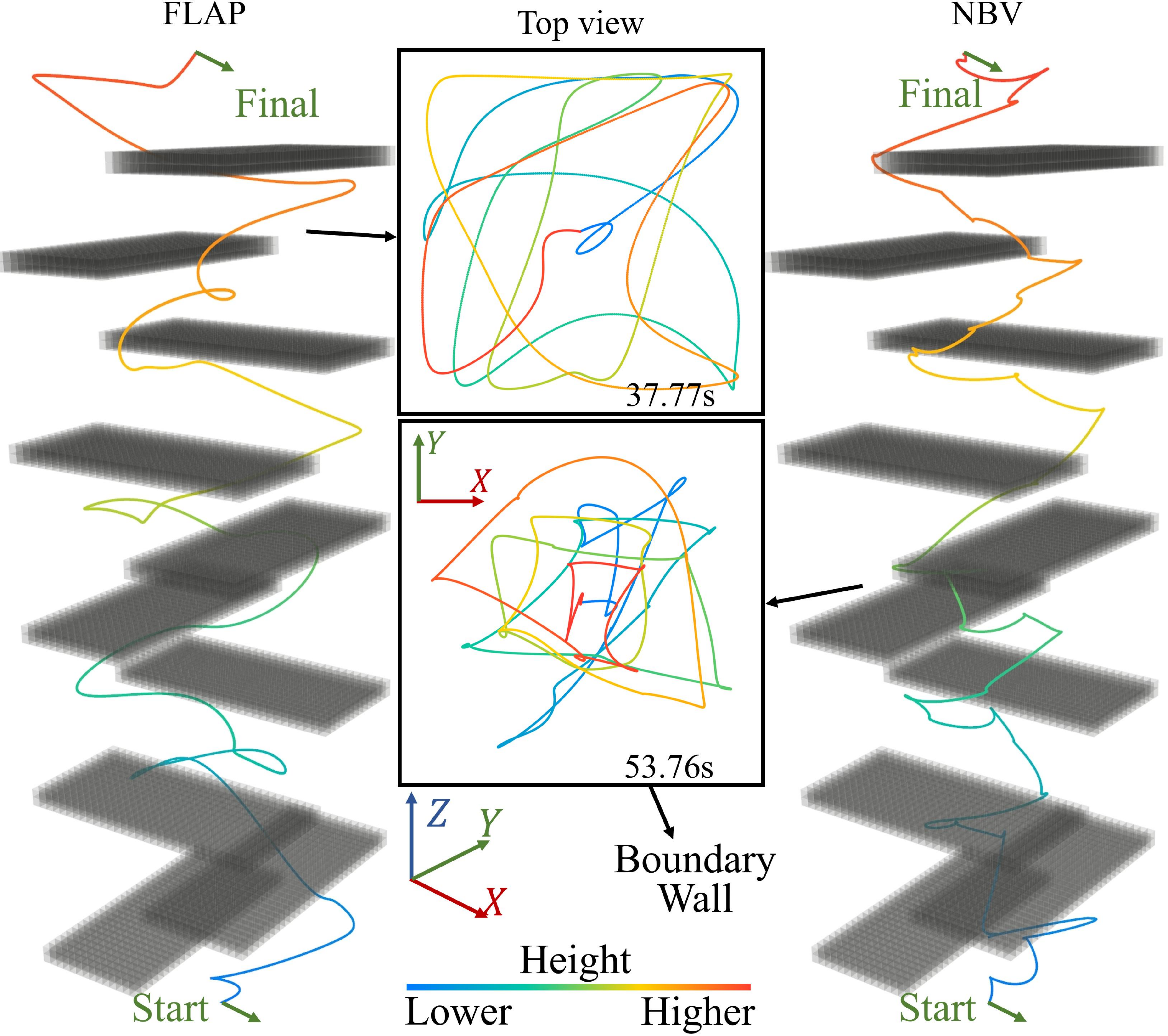}
  \caption{
      Performance of \ACRONYM{} and NBV in a cluttered environment with overlapping vertical obstacles. 
      The UAV, equipped with a vision camera, must traverse a vertically constrained passage formed by these obstacles. 
      The trajectory is color-coded by height. 
      The left and right figures show the results of \ACRONYM{} and NBV, respectively. 
      For visualization clarity, the pipeline walls are omitted in the figure to reveal the internal obstacle layout and trajectories.
      The center top view compares the two trajectories and indicates the obstacle boundaries and the travel time from start to goal.
    }
  \label{fig:camera_pipeline}
  \vspace{-0.4cm}
\end{figure}

To further validate the effectiveness of \ACRONYM{} in vertical environments, we design a narrow vertical pipeline scenario, as shown in Fig.~\ref{fig:camera_pipeline}. 
The pipeline walls contain several flat obstacles, each with a width equal to half the pipeline's side length, spaced $1\,\mathrm{m}$ apart.
The UAV must take off from the bottom of the pipeline, navigate through multiple obstacles, and reach the target at the top of the pipeline.
The UAV's maximum velocity is set to $2\,\mathrm{m/s}$.

As shown in Fig.~\ref{fig:camera_pipeline}, among the evaluated methods, only \ACRONYM{} and NBV successfully complete the task in this vertically constrained environment.
NBV exhibits noticeable pauses upon reaching target viewpoints, requiring the UAV to frequently accelerate and decelerate, which prolongs the overall flight time.
In contrast, \ACRONYM{} plans a continuous and smooth trajectory using its active perception strategy.
To maintain efficient observation of the unknown space, \ACRONYM{} often flies closer to the outer walls of the pipeline, thereby improving overall operational efficiency.

\section{Real-World Experiments}\label{sec:real_world}
We validate \ACRONYM{} in several representative real-world environments.

\begin{figure}[b]
  \centering
  \vspace{-0.3cm}
  \includegraphics[width=\linewidth]{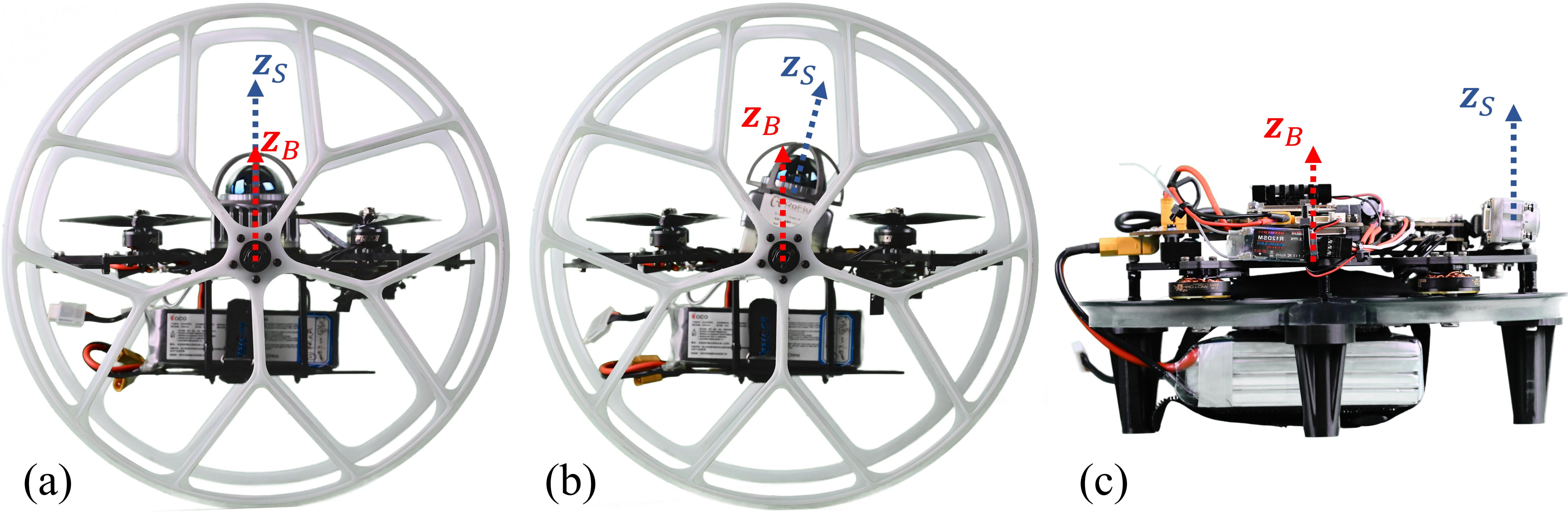}
  \caption{
    UAV platforms with three sensor configurations: (a) horizontal LiDAR, (b) inclined LiDAR, and (c) camera. 
    The z-axes of the body frame $B$ and the sensor frame $S$ are shown. 
    The LiDAR-equipped UAVs are fitted with passive wheels as a protective mechanism during safety tests.
  }
  \label{fig:real_uavs}
\end{figure}

\subsection{Traverse Tall Obstacles}

This experiment evaluates the planner's ability to use active perception to traverse tall vertical obstacles.
The UAV must climb over a high wall and then execute a cautious descent into the area with obstacles on the other side, rather than descending aggressively without sufficient observation. 
This showcases the planner's ability to handle complex vertical maneuvers, rather than merely performing horizontal obstacle avoidance. 
To evaluate the planner's performance under different sensing modalities, we implement three distinct sensor configurations:

\begin{itemize}
  \item Horizontal LiDAR: A Mid-360 LiDAR is mounted on the UAV with its z-axis parallel to the UAV's z-axis, as shown in Fig.~\ref{fig:real_uavs}(a).
  \item Inclined LiDAR: A Mid-360 LiDAR is tilted forward by $15^\circ$ about the UAV's y-axis, a common setup that better exploits the sensor FOV, as shown in Fig.~\ref{fig:real_uavs}(b). 
  \item Camera Sensor: The UAV carries an Orbbec Gemini 335 depth camera with its optical axis aligned to the UAV forward ($x$) direction, as shown in Fig.~\ref{fig:real_uavs}(c). 
\end{itemize}

\begin{figure*}[t]
  \centering
  \includegraphics[width=\textwidth]{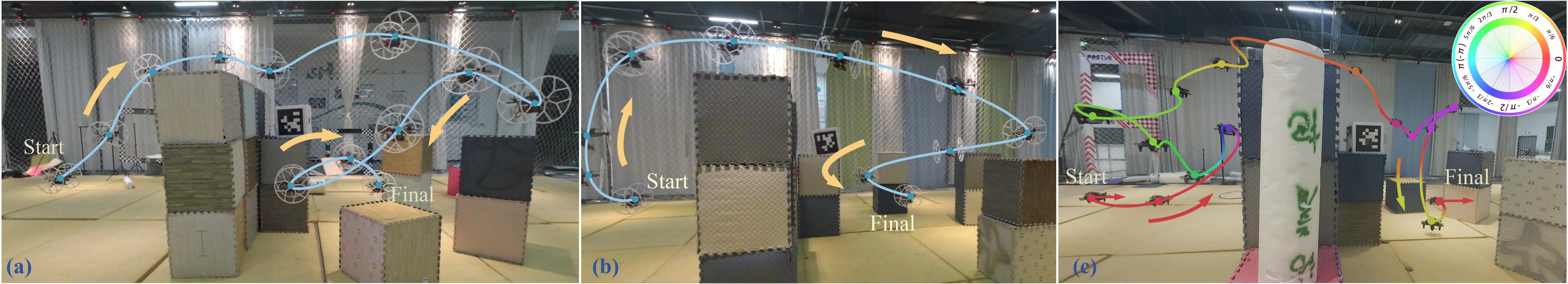}
  \caption{
    Experimental results with three UAV sensing configurations: (a) horizontal LiDAR, (b) inclined LiDAR, and (c) camera.
    The UAV must rely on onboard sensing to traverse a tall frontal obstacle and land on the far side. 
    For the vision case, the UAV's yaw angle along the trajectory is encoded by gradient colors.
    }
  \label{fig:real_three_uav}
  \vspace{-0.6cm}
\end{figure*}

With the Livox Mid-360 LiDAR, the horizontal FOV is $360^\circ$, while the vertical FOV extends upward by about $52^\circ$ and downward by only about $7^\circ$. 
The relatively large upward FOV allows the UAV to easily observe and climb over the frontal obstacle. 
After crossing it, however, the much smaller downward FOV makes it difficult to keep the area below in view during descent.
To maintain observability, the UAV has to pull back from the region of interest before descending further, resulting in a wave-like flight path, as shown in Fig.~\ref{fig:real_three_uav}(a).
This is due to the UAV's size, as regions within the FOV are not necessarily safe, and obstacles in blind regions may also pose collision risks.
Consequently, the UAV climbs (e.g., at the far right of Fig.~\ref{fig:real_three_uav}(a)) to avoid the obstacle in blind zones and similarly performs an upward maneuver near the goal.
In most works using the Mid-360, the LiDAR is tilted to balance the vertical FOV, yielding approximately $[-22^\circ, 37^\circ]$ in the forward direction. 
The upper bound is set slightly higher to account for forward pitch during flight. 
Our experiments confirm the effectiveness of this configuration, as shown in Fig.~\ref{fig:real_three_uav}(b). 
When flying over the obstacle, the UAV first approaches laterally and yaws to expose its side toward the goal.
This is because the tilted LiDAR provides a larger viewing angle to the UAV's side.
After crossing, the UAV adjusts its orientation to observe lower spaces, enabling a faster approach to the goal. 
In contrast, the depth camera offers only about $[-45^\circ, 45^\circ]$ horizontal and $[-32^\circ, 32^\circ]$ vertical FOV, leading to more conservative motion; see Fig.~\ref{fig:real_three_uav}(c). 
The UAV follows a zigzag trajectory to climb into higher known free space while creating time for yaw adjustment.
During descent, acceleration induces a nose-down pitch that slightly improves observation efficiency, allowing the UAV to reach the goal with fewer zigzags.

\subsection{Horizontal and Vertical Active Perception}

\begin{figure}[t]
  \centering
  \includegraphics[width=\linewidth]{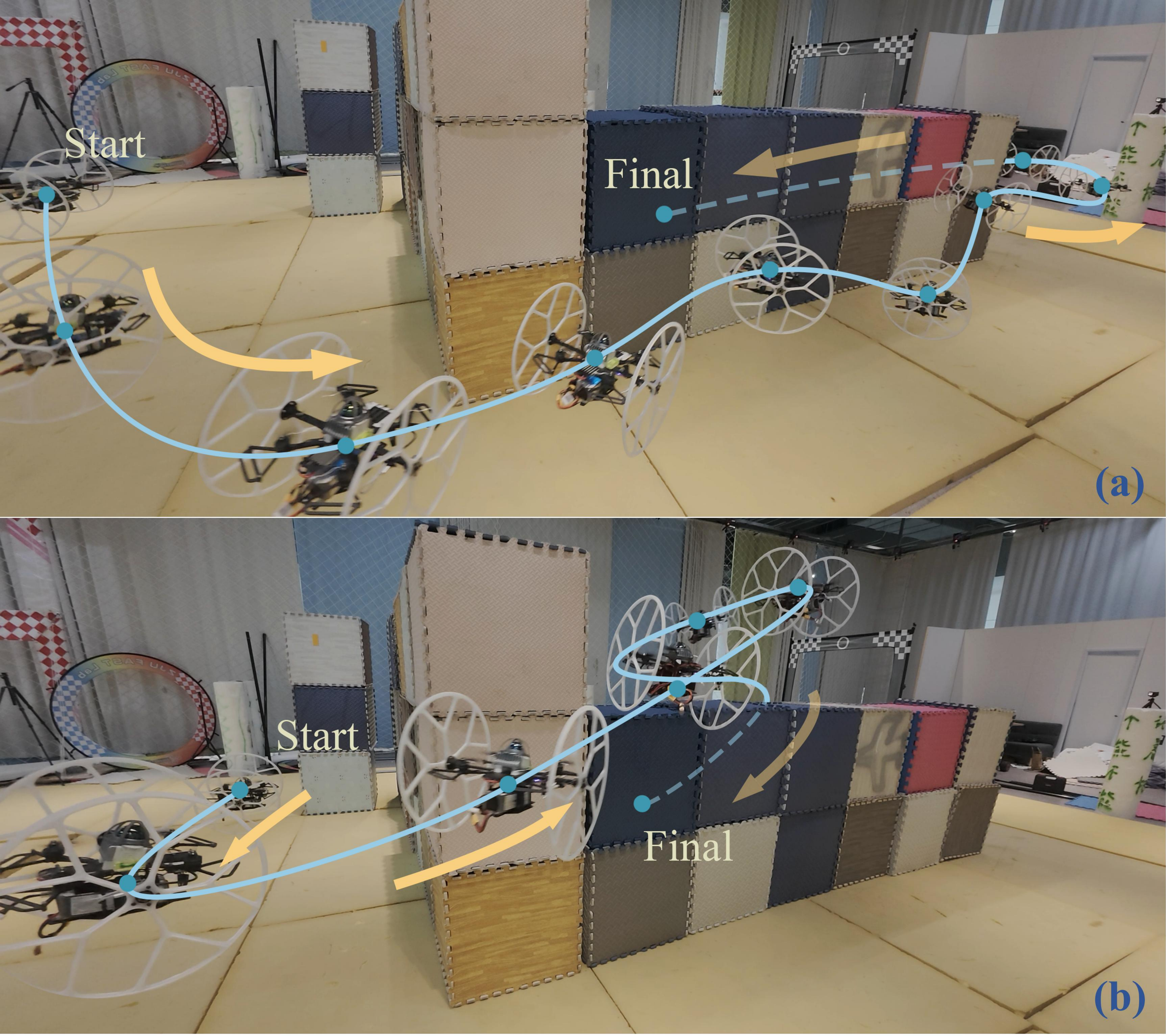}
  \caption{
    Planning results for the UAV equipped with a LiDAR sensor under height-restricted (a) and unrestricted (b) vertical motion, in an environment with a tall frontal obstacle and a lower obstacle on the right.
  }
  \label{fig:real_radar_hori}
  \vspace{-0.3cm}
\end{figure}

This experiment shows how \ACRONYM{} exploits vertical motion for active perception, rather than being limited to horizontal detours.
We design two distinct strategies.
In the vertically restricted strategy, which is commonly used for UAVs with limited vertical FOV, the UAV is constrained to search for feasible paths mainly in the horizontal direction. 
This reduces the risk of colliding with obstacles in vertical blind regions. 
In the vertically unrestricted strategy, the UAV is allowed to move along the z-axis to seek feasible observation and traversable space.

As shown in Fig.~\ref{fig:real_radar_hori}, we test the UAV equipped with an inclined LiDAR in an environment with a tall obstacle ahead and a slightly lower obstacle to the right.
Under the vertically restricted strategy shown in Fig.~\ref{fig:real_radar_hori}(a), the UAV is forced to take a horizontal detour.
In contrast, \ACRONYM{} shown in Fig.~\ref{fig:real_radar_hori}(b) exploits the vertical dimension and passes over the obstacle. 
Due to the LiDAR's inclined installation, which offers a downward view mainly forward, the UAV actively adjusts its yaw during descent to improve the observation of unexplored areas.

\begin{figure}[t]
  \centering
  \includegraphics[width=\linewidth]{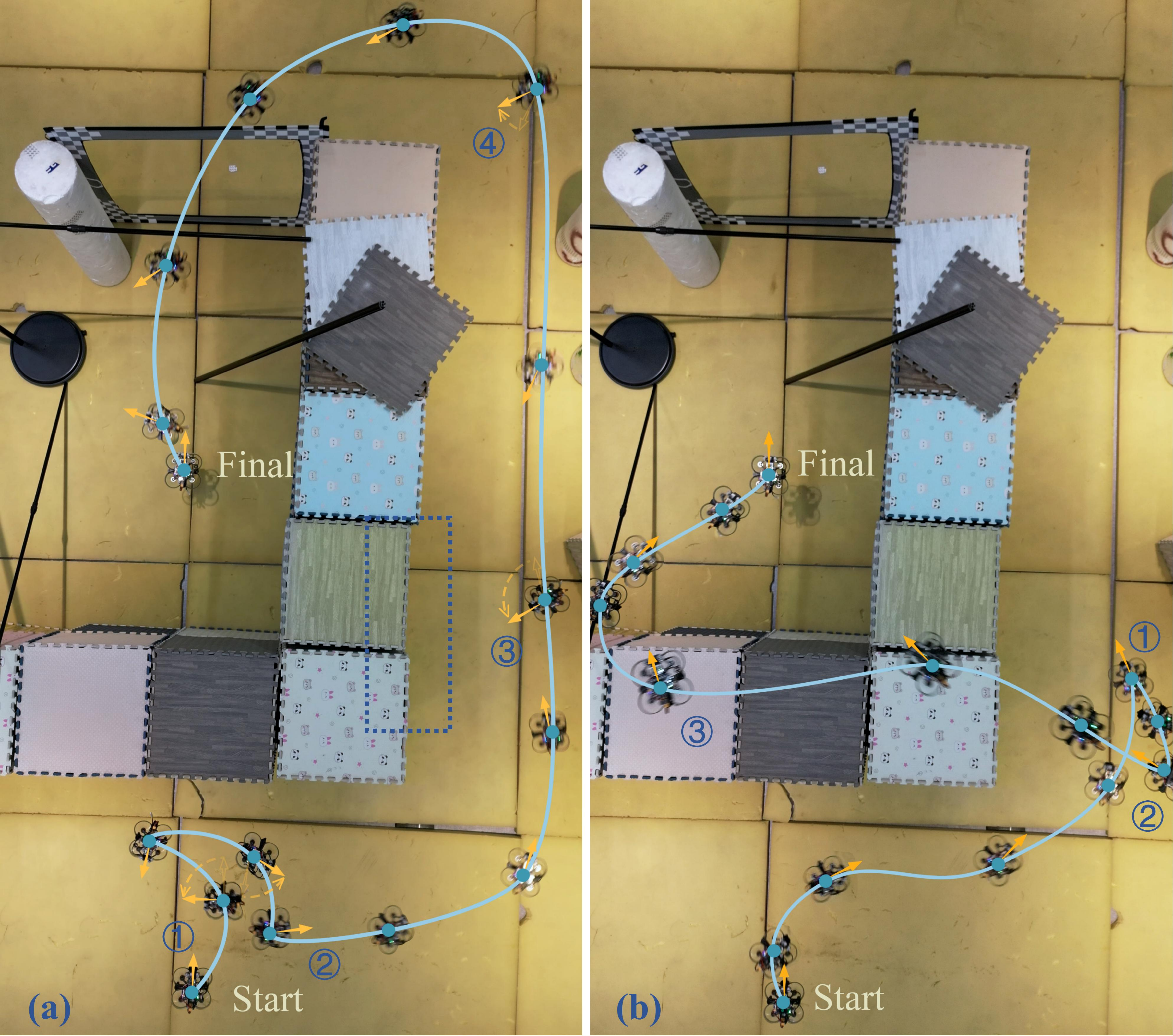}
  \caption{
    Planning results for the UAV equipped with a camera sensor with restricted (a) and unrestricted (b) vertical motion in an L-shaped obstacle environment.
  }
  \label{fig:real_camera_hori}
  \vspace{-0.2cm}
\end{figure}

We test the camera-equipped UAV in an environment with an L-shaped obstacle, as shown in Fig.~\ref{fig:real_camera_hori}.
We first restrict vertical motion to evaluate \ACRONYM{} in a horizontally occluded environment, as shown in Fig.~\ref{fig:real_camera_hori}(a). 
Upon takeoff, the UAV prioritizes exploring the left side due to initial environmental uncertainty at \ding{172}.
Upon finding the left side impassable, it diverts to the right at \ding{173}.
Guided by the front-end path, the planner adjusts its yaw angle at \ding{174} to inspect a possible shortcut occluded by the obstacle (dashed box).
After confirming the detour is necessary, it accelerates and observes the unknown space at \ding{175} before reaching the goal. 
As shown in Fig.~\ref{fig:real_camera_hori}(b), when vertical motion is permitted, the UAV ascends from \ding{172} to \ding{173} to observe the top of the obstacle instead of following a horizontal detour.
It then heads toward the goal at \ding{174}, following an observation strategy enabled by vertical motion.
Notably, in Fig.~\ref{fig:real_camera_hori}(b), the UAV does not inspect the left side.
One possible explanation is the high uncertainty in the initial planning state and in the early observations, which may bias the planner toward the right side. 
However, this behavior is not the focus of this experiment.
These results demonstrate that vertical active perception expands the planner's available observation strategies and enables it to handle occlusions in 3D space, rather than being restricted to horizontal detours.

\subsection{Outdoor Experiments}

\begin{figure}[t]
  \centering
  \includegraphics[width=\linewidth]{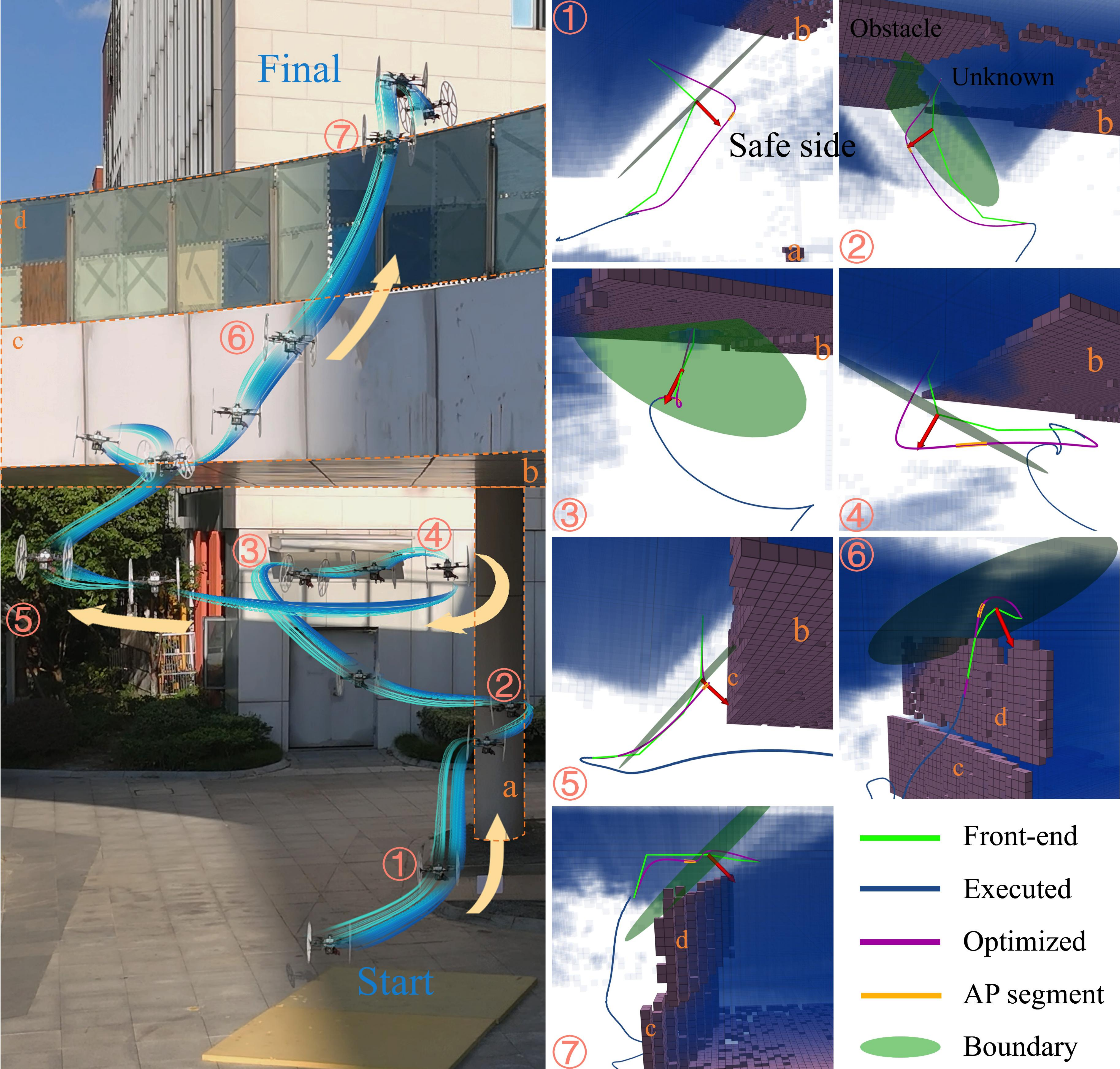}
  \caption{
    Planning results of \ACRONYM{} in an outdoor environment. 
    The left figure shows the complete trajectory, while the right figures present the corresponding trajectory and the map at representative instances. 
    Key obstacles are marked with lowercase letters to aid interpretation of the viewpoints.
  }
  \label{fig:real_radar_layer}
  \vspace{-0.4cm}
\end{figure}

This experiment showcases the planner's capability for 3D navigation in an unstructured outdoor environment. 
The objective is to move from the ground to an elevated gallery, demonstrating reliable obstacle avoidance and active perception in outdoor space. 

The UAV is equipped with a horizontal LiDAR, as shown in Fig.~\ref{fig:real_uavs}(a). 
Because of this limited vertical observability, part of the overhead space is initially outside the sensor's view. 
The UAV first explores upward until it confirms that the direct overhead route is blocked, as shown at \ding{172}--\ding{174} in Fig.~\ref{fig:real_radar_layer}. 
The planner then chooses a horizontal detour. 
While the UAV passes beneath wall $b$, the regions around walls $c$ and $d$ remain unmapped. 
It therefore increases its distance to these walls to obtain better visibility, as shown at \ding{175} and \ding{176} in Fig.~\ref{fig:real_radar_layer}. 
After identifying feasible free space near them, the UAV ascends to the goal height at \ding{177} and then descends to the final low-altitude goal position at \ding{178}.

\section{Conclusion}\label{sec:conclusion}
In this paper, we present a trajectory optimization framework that integrates active perception into UAV planning for 3D unknown environments with restricted sensor FOV. 
We formulate visibility and safety criteria in the sensor frame and introduce parameterized active perception segments with velocity-dependent activation functions. 
This approach promotes early exploration of unknown spaces while maintaining dynamic feasibility and computational efficiency.
Extensive simulations and real-world experiments show that the planner can generate smooth trajectories while maintaining safety and efficiency, and that it performs better than the baselines in the presented scenarios.
Since the known-unknown boundary is determined by the front-end, enhancing its quality will generate superior global paths that improve overall trajectory efficiency without substantial additional computation.
Additionally, we plan to extend the framework to multi-sensor fusion scenarios, exploiting complementary sensor modalities to enhance perception robustness and planning performance.

\bibliography{references}

\end{document}